\documentclass[man]{apa7}

\usepackage[american]{babel}

\usepackage{csquotes}
\usepackage[style=apa,sortcites=true,sorting=nyt,backend=biber]{biblatex}
\DeclareLanguageMapping{american}{american-apa}
\usepackage{booktabs}       
\usepackage{amsfonts}       
\usepackage{longtable}
\usepackage{multirow,setspace}
\usepackage{pdfpages}
\usepackage{tikz}
\usepackage{array}
\usepackage{amsmath}
\usepackage{amssymb}
\usepackage{mathtools}
\usepackage[flushleft]{threeparttable}
\usepackage{bm} 
\usepackage{bbm} 
\usepackage{siunitx} 
\usepackage{etoolbox} 
\usepackage{makecell} 
\usepackage{calc} 
\usepackage{float} 
\usetikzlibrary{positioning}
\usepackage{algorithm} 
\usepackage{url} 
\usepackage{subcaption} 
\usepackage{enumitem} 
\usepackage{amsthm}
\newtheorem{theorem}{Theorem}
\newtheorem*{remark}{Remark}

\DeclareUnicodeCharacter{0392}{ }

\robustify\bfseries

\DeclareMathOperator{\mse}{MSE}
\DeclareMathOperator*{\argmax}{arg\,max}
\DeclareMathOperator{\dkl}{\textit{D}_{KL}}
\DeclareMathOperator{\elbo}{ELBO}
\DeclareMathOperator{\iwelbo}{IW-ELBO}

\DeclareMathOperator{\vech}{vech}
\DeclareMathOperator{\diag}{diag}
\DeclareMathOperator{\bias}{bias}
\DeclareMathOperator{\train}{train}
\DeclareMathOperator{\test}{test}
\DeclareMathOperator{\acc}{\textit{acc}}
\DeclareMathOperator{\bernoulli}{Ber}
\DeclareMathOperator{\binomial}{Bin}
\DeclareMathOperator{\base}{base}
\DeclareMathOperator{\prop}{prop}
\DeclareMathOperator{\rfi}{C2ST-RFI}
\DeclareMathOperator{\plim}{plim}
\DeclareMathOperator{\power}{power}
\DeclareMathOperator{\imp}{\textit{importance}}

\makeatletter
\let\pgfmathrandomX=\pgfmathrandom@
\usepackage{pgfplots}%
\let\pgfmathrandom@=\pgfmathrandomX
\makeatother
\pgfplotsset{compat=1.14}
\usepgfplotslibrary{fillbetween}
\usepackage{ifthen}

\usetikzlibrary{shapes,arrows,positioning,calc,chains,scopes}

\definecolor{snowymint}{HTML}{E3F8D1}
\definecolor{wepeep}{HTML}{FAD2D2}
\definecolor{portafino}{HTML}{F5EE9D}
\definecolor{plum}{HTML}{DCACEF}
\definecolor{sail}{HTML}{A3CEEE}
\definecolor{highland}{HTML}{6D885A}

\tikzstyle{signal}=[arrows={-latex},draw=black,line width=1.5pt,rounded corners=4pt]

\tikzstyle{block}=[draw=black,line width=1.0pt]
\tikzstyle{cell}=[style=block,draw=highland,fill=snowymint,
    rounded corners]
\tikzstyle{celllayer}=[style=block,draw,fill=portafino,
    inner sep=1pt,outer sep=0,
    minimum width=28pt, minimum height=14pt]
\tikzstyle{pointwise}=[style=block,ellipse,fill=wepeep,
    inner sep=1pt,outer sep=0, minimum size=12pt]

\tikzstyle{netnode}=[circle, inner sep=0pt, text width=22pt, align=center, line width=1.0pt]
\tikzstyle{inputnode}=[netnode, fill=white,draw=black]
\tikzstyle{hiddennode}=[netnode, fill=white,draw=black]
\tikzstyle{outputnode}=[netnode, fill=white,draw=black]

\def\layerwidth{90pt}
\def\layerheight{14pt}

\tikzstyle{layer}=[style=block, draw, fill=black!20!white,
    inner sep=1pt,outer sep=0, font=\footnotesize,
    text centered, 
    minimum width=\layerwidth, minimum height=\layerheight]

\tikzstyle{fc}=[style=layer, fill=blue!30!white]
\tikzstyle{conv}=[style=layer, fill=green!30!white]
\tikzstyle{activation}=[style=layer, fill=orange!30!white]
\tikzstyle{pool}=[style=layer, fill=red!30!white]
\tikzstyle{bn}=[style=layer, fill=cyan!30!white]
\tikzstyle{recurrent}=[style=layer, fill=purple!30!white]
\tikzstyle{softmax}=[style=layer, fill=yellow!30!white]
\tikzstyle{point}=[]
\tikzstyle{branch}=[coordinate]

\def\vlayerwidth{30pt}
\def\vlayerheight{3pt}
\def\vblockheight{28pt}

\tikzstyle{vlayer}=[minimum width=\vlayerwidth, minimum height=\vlayerheight]
\tikzstyle{vblock}=[minimum width=\vlayerwidth, minimum height=\vblockheight, text width=1cm, align=center]

\colorlet{fn}{gray!90!green!30!white}
\colorlet{tp}{green!40!white}
\colorlet{fp}{red!40!white}
\colorlet{tn}{gray!90!red!20!white}

\title{Deep Learning-Based Estimation and Goodness-of-Fit for Large-Scale Confirmatory Item Factor Analysis}
\shorttitle{Deep Confirmatory IFA}

\author{Christopher J. Urban and Daniel J. Bauer}

\affiliation{L. L. Thurstone Psychometric Laboratory in the Department of Psychology and Neuroscience, University of North Carolina at Chapel Hill}

\leftheader{ }

\authornote{This material is based upon work supported by the National Science Foundation Graduate Research Fellowship under Grant No. DGE-1650116. \\
Correspondence should be sent to cjurban@live.unc.edu.}

\abstract{We investigate novel parameter estimation and goodness-of-fit (GOF) assessment methods for large-scale confirmatory item factor analysis (IFA) with many respondents, items, and latent factors. For parameter estimation, we extend \citeauthor{UrbanBauer2021}'s (\citeyear{UrbanBauer2021}) deep learning algorithm for exploratory IFA to the confirmatory setting by showing how to handle constraints on loadings and factor correlations. For GOF assessment, we explore simulation-based tests and indices that extend the classifier two-sample test (C2ST), a method that tests whether a deep neural network can distinguish between observed data and synthetic data sampled from a fitted IFA model. Proposed extensions include a test of approximate fit wherein the user specifies what percentage of observed and synthetic data should be distinguishable as well as a relative fit index (RFI) that is similar in spirit to the RFIs used in structural equation modeling. Via simulation studies, we show that: (1) the confirmatory extension of \citeauthor{UrbanBauer2021}'s (\citeyear{UrbanBauer2021}) algorithm obtains comparable estimates to a state-of-the-art estimation procedure in less time; (2) C2ST-based GOF tests control the empirical type I error rate and detect when the latent dimensionality is misspecified; and (3) the sampling distribution of the C2ST-based RFI depends on the sample size.}


\makeatletter
\def\def@donotrepeattitle{} 
\makeatother

\begin{document}

\maketitle

\setcounter{secnumdepth}{3} 

\section{Introduction} \label{Intro}

Item factor analysis \parencite[IFA;][]{Bock1988} is an invaluable method for investigating the latent structure underlying the discrete item response data that arises in many social science applications. In particular, IFA allows researchers to summarize a large number of item responses using a smaller number of continuous latent factors, thereby reducing the dimensionality of the data and potentially making the data easier to understand. Researchers with specific hypotheses about the number of factors, the relations between the item responses and the factors, and the factor correlations typically encode their hypotheses as parameter constraints in a confirmatory IFA model \parencite[e.g.,][]{Wirth2007}. For example, personality assessments such as the revised Minnesota Multiphasic Personality Inventory \parencite{Butcher1989} and the International Personality Item Pool NEO \parencite{Goldberg1999} include hundreds of items organized into subscales wherein each subscale is designed to measure a single personality factor. Modeling such designs using confirmatory IFA entails estimating the relations between the items comprising a subscale and their corresponding factor while constraining these items' relations with all other factors to zero (i.e., not estimating these relations). Estimating confirmatory IFA model parameters and their standard errors permits inferences about the properties of items as well as about the characteristics of the population from which the observed sample was drawn. Subsequent goodness-of-fit (GOF) analyses provide useful information about how well the fitted model approximates the data generating model \parencite[e.g.,][]{Maydeu-Olivares2013}.

Unfortunately, both parameter estimation and GOF assessment have long been computationally challenging in the large-scale setting with many respondents, items, and latent factors \parencite[e.g.,][]{Cai2010, Cai2010a}. Existing methods for confirmatory IFA may therefore be sub-optimal for analyzing complex, high-dimensional item response data arising from sources such as surveys, standardized tests, online applications, and electronic data capture, many of which are becoming increasingly available to social scientists \parencite[e.g.,][]{Pardos2017, Woo2020}. To clarify this issue, consider \citeauthor{Bock1981}'s (\citeyear{Bock1981}) marginal maximum likelihood (MML) estimator, which has many desirable statistical properties and is typically the preferred estimator for confirmatory IFA parameter estimation.\footnote{We note that other estimators such as limited-information estimators \parencite[e.g.,][]{Joreskog2001, Muthen1978, Muthen1984} and joint maximum likelihood (JML) estimators \parencite{Chen2019} are more computationally efficient than the MML estimator. However, these alternative estimators have different statistical properties --- for example, limited-information estimators are not asymptotically efficient, while JML estimators are only consistent when the sample size and the number of items simultaneously tend to infinity --- and are not considered further here due to space constraints.} The MML approach bases inference on the marginal likelihood of the observed item responses, which is obtained by integrating out the latent factors. Problematically, however, evaluating this integral is computationally burdensome when the number of factors $P$ is even moderately large (e.g., $P \geq \num{5}$). Researchers have devised numerous methods to avoid this computational burden: adaptive Gaussian quadrature methods \parencite{Rabe-Hesketh2005, Schilling2005}, Laplace approximation methods \parencite[e.g.,][]{Huber2004}, Monte Carlo expectation-maximization (EM) algorithms \parencite[e.g.,][]{Meng1996, Song2005}, Markov Chain Monte Carlo methods \parencite[e.g.,][]{Beguin2001, Edwards2010}, and stochastic approximation methods \parencite[SA; e.g.,][]{Cai2010, Cai2010a, Zhang2020}. Of the above methods, SA procedures such as the Metropolis-Hastings Robbins-Monro (MH-RM) algorithm \parencite{Cai2010} and the stochastic EM (StEM) algorithm \parencite{Zhang2020} are the most computationally efficient; in recent years, MH-RM has been particularly widely used in the social and behavioral sciences due to its flexibility and computational efficiency. However, even these state-of-the-art SA procedures are slow when the sample size $N$, number of items $J$, and number of factors $P$ are all large (e.g., $N \geq \num{10000}$, $J \geq \num{100}$, and $P \geq \num{10}$). 

Even after surmounting the computational difficulties associated with MML estimation, researchers who wish to assess their confirmatory IFA model's GOF face yet another computational barrier. Let $K_j$ denote the number of response categories for item $j$. To simplify the presentation, assume all items have the same number of categories such that $K = K_j$ for $j = 1, \ldots, J$. GOF assessment for IFA models fitted via MML is typically based on the underlying $K^J$-dimensional multinomial table on which the model is defined. Full-information GOF statistics such as Pearson's statistic and the likelihood ratio statistic have inaccurate $p$-values when the number of items and the number of response categories are even moderately large due to the multinomial table's sparseness \parencite[e.g., when $J \geq 6$  and $K \geq 5$;][]{Thissen1997}. Limited-information GOF statistics overcome the sparsity problem by only utilizing marginals of the multinomial table, thereby ``concentrating'' the information available for testing to obtain more accurate \textit{p}-values and higher power \parencite[e.g.,][]{Maydeu-Olivares2005, Maydeu-Olivares2006, Maydeu-Olivares2014}. However, calculating limited-information GOF statistics requires high-dimensional numerical integration and is typically computationally intensive. Despite efforts to improve computational efficiency for specific IFA models \parencite[e.g.,][]{Cai2013}, calculating limited-information GOF statistics for general  confirmatory IFA models remains computationally intensive when the number of items, response categories, and factors are all large (e.g., $J \geq 100$, $K \geq 5$, and $P \geq 10$).

Based on the preceding discussion, it is clear that more computationally efficient MML estimation and GOF assessment methods are needed to apply confirmatory IFA to very large-scale data. In this work, we investigate deep learning methods that offer steps toward addressing some of the difficulties mentioned above. The method we propose for parameter estimation is based on \citeauthor{UrbanBauer2021}'s (\citeyear{UrbanBauer2021}) deep learning algorithm for exploratory IFA. Their algorithm uses an importance-weighted amortized variational estimator (I-WAVE) that combines variational inference and importance sampling to construct an approximation to the MML estimator. By increasing the number of importance-weighted samples drawn during fitting, the I-WAVE typically trades computational efficiency for a better approximation. In the large-scale exploratory setting (i.e., $P = 10$, $J = 100$, $\num{1000} \leq N \leq \num{10000}$), the I-WAVE has empirically demonstrated comparable parameter estimation accuracy and increased computational efficiency relative to the MML estimator implemented via MH-RM \parencite{UrbanBauer2021}.

Assessing GOF for large-scale IFA models fitted via I-WAVE is not straightforward. In addition to being computationally inefficient in the large-scale setting, the limited-information GOF statistics described above were designed for models fitted via the MML estimator and have unknown theoretical properties when applied to models fitted via approximate MML (e.g., the I-WAVE). We aim to address both computational and theoretical issues simultaneously by instead considering simulation-based GOF assessment methods in which model-data fit is assessed by comparing the observed item responses to synthetic item responses sampled from a fitted IFA model. Previous work in simulation-based GOF assessment for confirmatory IFA has mainly focused on posterior predictive model checking (PPMC) in the Bayesian setting \parencite[e.g.,][]{Levy2009, Sinharay2006}. In PPMC, synthetic data simulated from the posterior predictive distribution are compared to the observed data using discrepancy measures (i.e., measures that quantify how two data sets differ). Unlike limited-information GOF statistics, PPMC is computationally efficient and does not rely on asymptotic arguments. Although PPMC was originally developed for Bayesian IFA models, recent work has shown that PPMC may be applied to frequentist IFA models using a normal approximation to the posterior predictive distribution \parencite{LeePPMC2016, Kuhfeld2019}. Unfortunately, PPMC assuming posterior normality (PPMC-N) is only well-motivated when models are fitted via exact maximum likelihood, suggesting that PPMC-N may not be a well-motivated GOF assessment method for I-WAVE.

The alternative GOF assessment methods we consider are based on a class of deep learning methods called classifier two-sample tests \parencite[C2STs;][]{Lopez-Paz2017}. Similar to $t$-tests and other classical two-sample tests, C2STs aim to determine whether two samples are drawn from the same distribution. To assess model-data fit in confirmatory IFA, C2STs begin by first sampling synthetic item responses from the fitted IFA model.\footnote{Unlike PPMC-N, C2STs are well-motivated when applied to an IFA model fitted via any estimator given that synthetic data can be sampled from the model. This holds for estimators that treat the latent factors as random effects (e.g., the MML estimator or the I-WAVE) but not for estimators that treat the latent factors as fixed effects (e.g., JML estimators).} The synthetic responses are combined with the observed responses to construct a new data set, which is divided at random into two disjoint subsets called the training set and the test set. Next, a deep neural network (NN) classifier is trained to distinguish between the observed and synthetic training set response patterns. Finally, the NN's test set accuracy (i.e., the proportion of test set response patterns correctly classified as observed or synthetic) is treated as the test statistic and is used to test the null hypothesis that the observed and synthetic data are drawn from the same distribution. For a perfect-fitting model, the accuracy obtained should not be significantly better than chance because the synthetic data will mimic the observed data characteristics. C2STs are computationally efficient in the large-scale setting, particularly when the NN is fitted using a scalable stochastic gradient method. When combined with variable importance measures \parencite{BreimanRF2001, Rudin2019}, C2STs also provide a variety of interpretable values to complement the use of $p$-values and may be viewed as uniting overall model fit, piece-wise (i.e., item-level) fit, and person fit under a single framework.

In exploring these issues, we make four primary contributions. First, we extend the I-WAVE to the confirmatory setting by showing how to handle user-defined constraints on the factor loadings and inter-factor correlations. Second, we empirically investigate confirmatory I-WAVE's finite sample behavior and conduct comparisons with MH-RM. Third, in addition to C2ST-based tests of perfect (exact) fit, we propose and explore a novel C2ST-based test of approximate fit as well as a C2ST-based relative fit index that is similar in spirit to the relative fit indices used in linear confirmatory factor analysis and structural equation modeling \parencite[SEM; e.g.,][]{Bollen1989, Bentler1980a, Tucker1973, Bentler1990}. Compared to standard C2STs, these new methods are potentially better suited to applications where the specified IFA model is unlikely to exactly capture the data generating model. Fourth, we conduct simulations to investigate the C2ST variants' finite sample behavior in several settings including when the IFA model is correctly and incorrectly specified.

The remainder of this paper is organized as follows. We begin with a review of variational methods (e.g., I-WAVE) for fitting confirmatory IFA models with polytomous responses. We next develop novel C2ST-based approximate GOF assessment methods. After providing implementation details, we investigate the proposed parameter estimation and GOF assessment methods' performance via an empirical example and simulation studies. We conclude by discussing limitations and extensions of the proposed methods.

\section{Variational Methods for Parameter Estimation}

Variational inference (VI) is an approach to approximate maximum likelihood estimation for latent variable (LV) models that is widely used in machine learning \parencite{Blei2017, Zhang2019}. VI has recently been applied for IFA in a variety of settings \parencite[e.g.,][]{Curi2019, Cho2020, ChenBeta32019, Natesan2016, Wu2020, Hui2017, UrbanBauer2021}. In this section, we review variational methods for IFA. We focus in particular on the VI-based method described by \textcite{UrbanBauer2021}, which provides a theoretical link between VI and MML estimation.

\subsection{A Model for Confirmatory IFA}

We first establish notation for the IFA models under consideration. Specifically, we consider \citeauthor{Samejima1969}'s (\citeyear{Samejima1969}) graded response model (GRM) for polytomous item responses, although the methods discussed in this section readily apply to other IFA models. Suppose that $N$ respondents have answered $J$ items. Let $x_{i,j} \in \{0, 1, \ldots, K_j - 1\}$ denote the response for respondent $i$ to item $j$ in $K_j$ ordinal categories. To simplify the presentation, assume $K_j = K$ for $j = 1, \ldots, J$. Note that when $K = 2$, the GRM reduces to the widely used multidimensional two-parameter logistic model \parencite{McKinley1983}.

Each respondent is represented by a $P \times 1$ latent vector $\mathbf{z}_i$ and each item is represented by a $(P + K - 1) \times 1$ parameter vector $\boldsymbol{\theta}_j = (\boldsymbol{\alpha}_j^\top, \boldsymbol{\beta}_j^\top)^\top$ where $\boldsymbol{\beta}_j$ is a $P \times 1$ vector of loadings and $\boldsymbol{\alpha}_j = (\alpha_{j1}, \ldots, \alpha_{j, K - 1})^\top$ is a $(K - 1) \times 1$ vector of strictly ordered category intercepts. The GRM defines a set of boundary response probabilities conditional on $\boldsymbol{\theta}_j$ and $\mathbf{z}_i$:
\begin{equation} \label{eq:2.1}
    \Pr(x_{i,j} \geq k \mid \boldsymbol{\theta}_j, \mathbf{z}_i) =
    \sigma\left[ \alpha_{j,k} + \boldsymbol{\beta}_j^\top \mathbf{z}_i \right], \quad k \in \{1, \ldots, K - 1\},
\end{equation}
where $\sigma[\cdot] = 1 / (1 + \exp[\cdot])$, $\Pr(x_{i,j} \geq 0 \mid \boldsymbol{\theta}_j, \mathbf{z}_i) = 1$, and $\Pr(x_{i,j} \geq K \mid \boldsymbol{\theta}_j, \mathbf{z}_i) = 0$. The conditional probability of the response $x_{i,j} = k$ for $k \in \{ 0, \ldots K - 1\}$ is
\begin{equation} \label{eq:2.2}
    \pi_{i,j,k} = \Pr(x_{i,j} = k \mid \boldsymbol{\theta}_j, \mathbf{z}_i) = \Pr(x_{i,j} \geq k \mid \boldsymbol{\theta}_j, \mathbf{z}_i) - \Pr(x_{i,j} \geq k + 1 \mid \boldsymbol{\theta}_j, \mathbf{z}_i).
\end{equation}

It follows from Equation (\ref{eq:2.2}) that the conditional distribution of $x_{i,j}$ is multinomial with $K$ cells, trial size $1$, and cell probabilities $\pi_{i,j,k}$:
\begin{equation} \label{eq:2.3}
    p_{\boldsymbol{\theta}_j}(x_{i,j}\mid \mathbf{z}_i) = \prod_{k = 0}^{K - 1} \pi_{i,j,k}^{\mathbbm{1}(x_{i,j} = k)},
\end{equation}
where $\mathbbm{1}(\cdot)$ denotes the indicator function. Let $\mathbf{x}_i = (x_{i,1}, \ldots, x_{i,n})^\top$ be the $i^\text{th}$ respondent's response pattern and let $\boldsymbol{\theta} = (\boldsymbol{\theta}_1^\top, \ldots, \boldsymbol{\theta}_J^\top)^\top$ be a vector collecting all item parameters. By the local independence assumption, the conditional distribution of $\mathbf{x}_i$ is
\begin{equation} \label{eq:2.4}
    p_{\boldsymbol{\theta}}(\mathbf{x}_i \mid \mathbf{z}_i) = \prod_{j = 1}^J p_{\boldsymbol{\theta}_j}(x_{i,j}\mid \mathbf{z}_i).
\end{equation}
Assume that $\mathbf{z}_i$ is multivariate normally distributed with zero mean vector and covariance matrix $\boldsymbol{\Sigma} = (\sigma_{p, p'})_{P \times P}$. Let $\boldsymbol{\omega} = (\boldsymbol{\theta}^\top, \vech(\boldsymbol{\Sigma})^\top)^\top$ be a vector collecting all unknown parameters. Under our assumptions about the distribution of the factors, the marginal distribution of $\mathbf{x}_i$ is given by
\begin{equation} \label{eq:2.5}
    p_{\boldsymbol{\omega}}(\mathbf{x}_i) = \int \prod_{j = 1}^J p_{\boldsymbol{\theta}_j}(x_{i,j}\mid \mathbf{z}) \mathcal{N}( \mathbf{z} \mid \boldsymbol{\Sigma} ) d\mathbf{z} ,
\end{equation}
where $\mathcal{N}(\cdot \mid \boldsymbol{\Sigma})$ is a normal density parameterized by $\boldsymbol{\Sigma}$ and the above integral is over $\mathbb{R}^P$.

We set $\sigma_{p, p} = 1$ for $p = 1, \ldots, P$ to identify the scale of the factors. In the confirmatory setting, users encode hypotheses about the measurement structure by placing restrictions on the loadings (e.g., by fixing $\beta_{j,p}$ to zero if item $j$ is not hypothesized to measure factor $p$). Following \textcite{Cai2010}, we consider the case of linear equality constraints so that the loadings may be written as
\begin{equation}
    \boldsymbol{\beta}_j = \mathbf{b}_j + \mathbf{A}_j \boldsymbol{\beta}_j',
\end{equation}
where $\boldsymbol{\beta}_j$ is the restricted loadings vector, $\mathbf{b}_j$ is a $P \times 1$ vector of constants, $\mathbf{A}_j$ is a $P \times P$ matrix of constants that implements the linear constraints, and $\boldsymbol{\beta}_j'$ is a vector of free parameters. \textcite{Cai2010} provides examples of how $\mathbf{b}_j$ and $\mathbf{A}_j$ may be specified to implement various restrictions, while \textcite{Anderson1956} provide sufficient conditions enabling $\mathbf{b}_j$ and $\mathbf{A}_j$ to be specified such that the model is identified.

\subsection{VI and Amortized VI}

Let $\mathbf{X}$ be an $N \times J$ matrix whose $i^\mathrm{th}$ row is $\mathbf{x}_i^\top$. The marginal log-likelihood (i.e., the evidence) of the observed data is
\begin{equation} \label{eq:2.6}
    \ell(\boldsymbol{\omega} \mid \mathbf{X}) = \sum_{i = 1}^N \log \bigg[ \int \prod_{j = 1}^J p_{\boldsymbol{\theta}_j}(x_{i,j}\mid \mathbf{z}_i) \mathcal{N}( \mathbf{z} \mid \boldsymbol{\Sigma} ) d\mathbf{z} ) \bigg].
\end{equation}
Maximizing $\ell(\boldsymbol{\omega} \mid \mathbf{X})$ by directly evaluating the $N$ integrals in Equation (\ref{eq:2.6}) is computationally intensive when $P$ is large. VI solves this issue by instead maximizing a computationally tractable lower bound on $\ell(\boldsymbol{\omega} \mid \mathbf{X})$. The evidence lower bound (ELBO) for a single observation is given by:
\begin{equation}
    \log p_{\boldsymbol{\omega}}(\mathbf{x}_i) \geq \elbo_i = \mathbb{E}_{q_{\boldsymbol{\psi}_i}(\mathbf{z}_i)}\big[ \log p_{\boldsymbol{\omega}}(\mathbf{x}_i \mid \mathbf{z}_i) \big] - \dkl \big[q_{\boldsymbol{\psi}_i}(\mathbf{z}_i) \| \mathcal{N}(\mathbf{z}_i \mid \boldsymbol{\Sigma})\big], \label{eq:2.10}
\end{equation}
where $\dkl \big[ \cdot \| \cdot \big]$ denotes the Kullback-Leibler (KL) divergence and $q_{\boldsymbol{\psi}_i}(\mathbf{z}_i)$ is an arbitrary density with parameter vector $\boldsymbol{\psi}_i$. The variational estimator of the IFA model parameters $\boldsymbol{\omega}$ is obtained by maximizing the ELBO over all observations w.r.t. both $\boldsymbol{\omega}$ and $\boldsymbol{\psi}_i$, which is equivalent to minimizing the KL divergence between $q_{\boldsymbol{\psi}_i}(\mathbf{z}_i)$ and the posterior distribution of the latent factors $p_{\boldsymbol{\omega}}(\mathbf{z}_i \mid \mathbf{x}_i)$ \parencite[][Sect. 4.2]{UrbanBauer2021}. Intuitively, obtaining the variational estimator drives $q_{\boldsymbol{\psi}_i}(\mathbf{z}_i)$ to approximate the true LV posterior; we henceforth refer to $q_{\boldsymbol{\psi}_i}(\mathbf{z}_i)$ as the approximate LV posterior. Following previous work \parencite[e.g.,][]{Kingma2014, Hui2017, UrbanBauer2021}, we set the approximate LV posterior to a computationally tractable isotropic normal density:
\begin{equation}
    q_{\boldsymbol{\psi}_i}(\mathbf{z}_i) = \mathcal{N}(\mathbf{z}_i \mid \boldsymbol{\mu}_i, \boldsymbol{\sigma}^2_i\mathbf{I}_P \big), \label{eq:2.11}
\end{equation}
where $\boldsymbol{\mu}_i$ is a $P \times 1$ vector of means, $\boldsymbol{\sigma}^2_i$ is a $P \times 1$ vector of variances, and $\mathbf{I}_P$ is a $P \times P$ identity matrix.

Although traditional VI fits a different parameter vector $\boldsymbol{\psi}_i$ (i.e., a different mean $\boldsymbol{\mu}_i$ and variance $\boldsymbol{\sigma}_i^2$) for each observation, this approach quickly becomes computationally infeasible for large sample sizes. Amortized variational inference (AVI) solves this issue by parameterizing the approximate posterior using a powerful function approximator called an inference model. Since the inference model parameters are shared across observations, AVI fits a constant number of parameters regardless of the sample size, whereas VI fits a number of parameters that grows with the sample size.

The variational autoencoder \parencite[VAE; ][]{Kingma2014, Rezende2014} is an AVI algorithm that uses a deep NN inference model \parencite[for a brief overview of NNs, see][Sect. 2]{UrbanBauer2021}. Using an NN inference model is considered justifiable because NNs can approximate any Borel measurable function \parencite[e.g.,][]{Cybenko1989} and perform well in real-world applications \parencite{LeCun2015}. We can specify a VAE for confirmatory IFA by parameterizing the approximate LV posterior as follows:
\begin{align} \label{eq:2.12}
    \begin{split}
        ( \boldsymbol{\mu}_i^\top, \boldsymbol{\sigma}_i^\top)^\top &= f_{\boldsymbol{\psi}}(\mathbf{x}_i), \\
        q_{\boldsymbol{\psi}}(\mathbf{z}_i \mid \mathbf{x}_i) &= \mathcal{N} (\mathbf{z}_i \mid \boldsymbol{\mu}_i, \boldsymbol{\sigma}^2_i\mathbf{I}_P ),
    \end{split}
\end{align}
where $\boldsymbol{\mu}_i$ is a $P \times 1$ predicted vector of means, $\boldsymbol{\sigma}_i$ is a strictly positive $P \times 1$ predicted vector of standard deviations, and $f_{\boldsymbol{\psi}}$ is an NN parameterized by $\boldsymbol{\psi}$. Instead of estimating a parameter vector $\boldsymbol{\psi}_i$ for each observation, the NN parameters $\boldsymbol{\psi}$ are now shared across observations.

\subsection{Importance-Weighted VI} \label{IWVI}

Importance-weighted VI \parencite{Burda2016, Domke2018} is a strategy for obtaining a better approximation to the true log-likelihood by increasing the flexibility of traditional VI. The importance-weighted amortized variational estimator (I-WAVE) for the IFA model parameters $\boldsymbol{\omega}$ is obtained by maximizing a new lower bound called the importance-weighted ELBO (IW-ELBO):
\begin{equation}
    \log p_{\boldsymbol{\omega}}(\mathbf{x}) \geq \iwelbo = \mathbb{E}_{\mathbf{z}_{1:R}}\bigg[ \log \frac{1}{R} \sum_{r = 1}^R w_r \bigg], \label{eq:2.18}
\end{equation}
where $\mathbf{z}_{1:R} \sim \prod_{r = 1}^R q_{\boldsymbol{\psi}}(\mathbf{z}_r \mid \mathbf{x})$, $w_r = p_{\boldsymbol{\omega}}(\mathbf{z}_r, \mathbf{x}) / q_{\boldsymbol{\psi}}(\mathbf{z}_r \mid \mathbf{x})$ are unnormalized importance weights for the joint distribution of latent and observed variables, $R$ is the number of importance-weighted (IW) samples, and we have dropped the case index $i$ for simplicity. The IW-ELBO reduces to the ELBO when $R = 1$ and converges monotonically to the marginal log-likelihood as $R \rightarrow \infty$ under mild assumptions \parencite{Burda2016}. This fact implies that I-WAVE and the MML estimator are equivalent when the number of IW samples $R$ equals infinity, in which case I-WAVE inherits the MML estimator's statistical properties.

Obtaining the I-WAVE requires an unbiased estimator for the gradient of the IW-ELBO w.r.t. $\boldsymbol{\xi} = (\boldsymbol{\omega}^\top, \boldsymbol{\psi}^\top)^\top$. Following \textcite{UrbanBauer2021}, we use \citeauthor{Burda2016}'s (\citeyear{Burda2016}) estimator for the IW-ELBO $\boldsymbol{\omega}$-gradient:
\begin{align}
    \nabla_{\boldsymbol{\omega}} \mathbb{E}_{\mathbf{x}_{1:R}}\bigg[ \log \frac{1}{R} \sum_{r = 1}^R w_r \bigg] &= \mathbb{E}_{\boldsymbol{\epsilon}_{1:R}} \bigg[ \sum_{r = 1}^R \widetilde{w}_r \nabla_{\boldsymbol{\omega}} \log w_r \bigg] \label{eq:2.19} \\
    &\approx \frac{1}{S} \sum_{s = 1}^S \bigg[ \sum_{r = 1}^R \widetilde{w}_{r, s} \nabla_{\boldsymbol{\omega}} \log w_{r, s} \bigg], \label{eq:2.20}
\end{align}
as well as \citeauthor{Tucker2019}'s (\citeyear{Tucker2019}) ``doubly reparameterized'' estimator for the IW-ELBO $\boldsymbol{\psi}$-gradient:
\begin{align}
    \nabla_{\boldsymbol{\psi}} \mathbb{E}_{\mathbf{x}_{1:R}}\bigg[ \log \frac{1}{R} \sum_{r = 1}^R w_r \bigg] &= \mathbb{E}_{\boldsymbol{\epsilon}_{1:R}}\bigg[ \sum_{r = 1}^R \widetilde{w}_r^2 \frac{\partial \log w_r}{\partial \mathbf{z}_r} \frac{\partial \mathbf{z}_r}{\partial \boldsymbol{\psi}} \bigg]^\top \label{eq:2.21} \\
    &\approx \frac{1}{S} \sum_{s = 1}^S \bigg[ \sum_{r = 1}^R \widetilde{w}_{r, s}^2 \frac{\partial \log w_{r, s}}{\partial \mathbf{z}_{r, s}} \frac{\partial \mathbf{z}_{r, s}}{\partial \boldsymbol{\psi}}\bigg]^\top \label{eq:2.22},
\end{align}
where (\ref{eq:2.20}) and (\ref{eq:2.22}) are Monte Carlo approximations to expectations (\ref{eq:2.19}) and (\ref{eq:2.22}), respectively; $\mathbf{z}_r = \boldsymbol{\mu} + \diag(\boldsymbol{\sigma}) \boldsymbol{\epsilon_r}$ with $\boldsymbol{\epsilon}_{1:R} \sim \prod_{r = 1}^R\mathcal{N}(\boldsymbol{\epsilon}_r)$; and $\widetilde{w}_r = w_r / \sum_{r' = 1}^R w_{r'}$ are normalized importance weights. Both estimators are unbiased, can be successfully approximated using a single Monte Carlo sample \parencite[as we do in this work;][]{Burda2016, Tucker2019}, and can be efficiently computed using an automatic differentiation procedure called backpropagation \parencite[e.g.,][]{Goodfellow2016}. After computing both estimators, we apply an adaptive stochastic gradient method called AMSGrad \parencite{Reddi2018} to iteratively update $\boldsymbol{\xi}$ until convergence.

\subsection{Handling User-Defined Constraints}

User-defined constraints on the factor loadings are straightforward to implement for I-WAVE. In particular, the gradient of the IW-ELBO w.r.t. the unconstrained loadings vector $\boldsymbol{\beta}_j'$ can be obtained using the chain rule:
\begin{equation} \label{eq:2.23}
   \nabla_{\boldsymbol{\beta}_j'} \iwelbo= \bigg( \frac{\partial \iwelbo}{\partial \boldsymbol{\beta}_j} \frac{\partial \boldsymbol{\beta}_j}{\partial \boldsymbol{\beta}_j'} \bigg)^\top = \mathbf{A}^\top_j \nabla_{\boldsymbol{\beta}_j} \iwelbo.
\end{equation}
Equation (\ref{eq:2.23}) implies that one can first compute the gradient of the IW-ELBO w.r.t. $\boldsymbol{\beta}_j$, then obtain the gradient w.r.t. $\boldsymbol{\beta}_j'$ via pre-multiplication by the transposed constraint matrix $\mathbf{A}^\top_j$.

Users also often wish to impose constraints on the factor correlation matrix $\boldsymbol{\Sigma}$. Let $\boldsymbol{\Sigma} = \mathbf{L} \mathbf{L}^\top$ where $\mathbf{L}$ is a $P \times P$ lower triangular matrix. We estimate $\mathbf{L}$ using a hyperspherical parameterization \parencite{Pinheiro1996, Rapisarda2007}, which enables unconstrained estimation of a variety of structured correlation matrices and has similar computational efficiency to estimating $\mathbf{L}$ directly. This parameterization is given by:
\begin{equation} \label{eq:2.24}
    l_{p, p'} =
        \begin{cases}
            \cos \vartheta_{p, 1}, & \text{if}\; p' = 1 \\
            \cos \vartheta_{p, p' + 1} \prod_{p'' = 1}^{p'} \sin \vartheta_{p, p''}, & \text{if}\; 1 < p' < p \\
            \prod_{p'' = 1}^{p} \sin \vartheta_{p, p''}, & \text{if}\; p' = p,
        \end{cases}
\end{equation}
for $p = 1, \ldots, P$ where $l_{p, p'}$ are elements of $\mathbf{L}$, $\vartheta_{1, 1} = \pi / 2$, and $\vartheta_{p, p'} \in (0, \pi]$ are angles measured in radians which are elements of a $P \times P$ lower triangular matrix $\boldsymbol{\Theta}$. Constraints on the angles giving rise to various correlation structures are discussed by \textcite{Tsay2017} as well as by \textcite{Ghosh2020}.

\section{Classifier Two-Sample Tests for Goodness-of-Fit Assessment}

\subsection{Exact C2STs} \label{ExactC2STs}

We now discuss the application of classifier two-sample tests \parencite[C2STs;][]{Lopez-Paz2017}, a class of simulation-based GOF assessment methods that have recently been developed in deep learning, to assessing exact GOF for confirmatory IFA models. Let $\hat{\boldsymbol{\omega}}$ denote parameter estimates obtained for some confirmatory IFA model. Let $\mathbf{x}_i \sim \mathbb{P}$ denote the $i^\text{th}$ observed response pattern and let $\mathbf{y}_j \sim  p_{\hat{\boldsymbol{\omega}}}(\mathbf{y}_j) = \hat{\mathbb{P}}$ denote the the $j^\text{th}$ synthetic response pattern drawn from the fitted model where $\mathbf{x}_i, \mathbf{y}_j \in \mathcal{X} = \bigtimes_{j = 1}^J \{ 0, \ldots, K - 1 \}$ for $i = 1, \ldots, N_1$ and $j = 1, \ldots, N_2$. To simplify the presentation, we assume $N_1 = N_2 = N$.

C2STs aim to test whether the observed and synthetic response patterns are drawn from the same distribution --- that is, C2STs aim to test $H_0: \mathbb{P} = \hat{\mathbb{P}}$ against $H_1: \mathbb{P} \neq \hat{\mathbb{P}}$. A C2ST is conducted by training an NN classifier to distinguish between the observed and synthetic response patterns. Intuitively, when $\mathbb{P} = \hat{\mathbb{P}}$, the NN's test set accuracy should be close to $1/2$ (i.e., chance), since samples from $\mathbb{P}$ and $\hat{\mathbb{P}}$ are indistinguishable. When $\mathbb{P} \neq \hat{\mathbb{P}}$, the NN should be able to capitalize on the distributional differences to obtain a test set accuracy higher than $1/2$. We now describe C2STs more formally as two-phase procedures consisting of a training phase and a testing phase.

The training phase begins with constructing a data set $D = \{ (\mathbf{x}_i, 1) \}_{i = 1}^{N} \cup \{ (\mathbf{y}_i, 0) \}_{i = 1}^{N} = \{ (\mathbf{u}_i, l_i) \}_{i = 1}^{2N}$. Next, $D$ is shuffled at random and split into disjoint sets $D = D_{\train} \cup D_{\test}$ where $N_{\train} = |D_{\train}|$ and $N_{\test} = |D_{\test}|$. Last, an NN classifier $f_{\boldsymbol{\phi}} : \mathcal{X} \rightarrow [0, 1]$ with parameters $\boldsymbol{\phi}$ is fitted by obtaining:
\begin{equation} \label{C2STObjective}
    \hat{\boldsymbol{\phi}} = \argmax_{\boldsymbol{\phi}} \sum_{i \in I_{\train}} \bigg[ l_i \log f_{\boldsymbol{\phi}}(\mathbf{u}_i) + (1 - l_i) \log \left( 1 -f_{\boldsymbol{\phi}}(\mathbf{u}_i) \right) \bigg],
\end{equation}
where $I_{\train} =\{ i : (\mathbf{u}_i, l_i) \in D_{\train} \}$. The objective in Equation (\ref{C2STObjective}) is just the log-likelihood for binary logistic regression. As with the IW-ELBO, we maximize the log-likelihood in (\ref{C2STObjective}) using the AMSGrad stochastic gradient method, which has guaranteed convergence to a stationary point under mild conditions \parencite{Zhou2018, Chen2019}.
    
The testing phase entails using the fitted NN $\hat{f} \coloneqq f_{\hat{\boldsymbol{\phi}}}$ to compute the test set classification accuracy:
\begin{equation} \label{eq:3.1}
    \widehat{\acc} = \frac{1}{N_{\test}} \sum_{i \in I_{\test}} \mathbbm{1} \left( \mathbbm{1} \left( \hat{f}(\mathbf{u}_i) > \frac{1}{2} \right) = l_i \right) = \frac{1}{N_{\test}} \sum_{ i \in I_{\test}} \widehat{\acc}_i,
\end{equation}
where $I_{\test} =\{ i : (\mathbf{u}_i, l_i) \in D_{\test} \}$. $\widehat{\acc}$ serves as our test statistic for deciding whether or not to reject $H_0$. For large $N_{\test}$, a $p$-value for $\widehat{\acc}$ can be obtained as follows:
\begin{equation} \label{eq:3.2}
    \hat{p} = \Pr(\widehat{\acc}' \geq \widehat{\acc} \mid H_0) \approx 1 - \Phi \left( \frac{\widehat{\acc} - 1/2}{\sqrt{\frac{1}{4 N_{\test}}}} \right),
\end{equation}
where $\Phi(\cdot)$ denotes the standard normal cumulative distribution function. To derive Equation (\ref{eq:3.2}), observe that $\acc_i \sim \bernoulli (\acc_i \mid p_i)$ where $p_i = 1/2$ is the probability of correctly classifying some $\mathbf{u}_i$ in the test set when $H_0$ is true. In this setting, Equation (\ref{eq:3.2}) follows from the fact that
\begin{equation} \label{eq:3.3}
    N_{\test}\acc \sim \binomial \left(N_{\test}\acc \,\middle\vert\, N_{\test}, \frac{1}{2} \right) \approx \mathcal{N} \left(N_{\test}\acc \,\middle\vert\, \frac{N_{\test}}{2}, \frac{N_{\test}}{4} \right)
\end{equation}
when $N_{\test}$ is large.

C2STs provide a variety of interpretable numbers that complement the use of $p$-values:
\begin{enumerate}[label={(\alph*)}]
    \item Taking $\hat{f}(\mathbf{u}_i)$ as an estimate of the conditional probability $\Pr(l_i = 1 \mid \mathbf{u}_i)$ for $i \in I_{\test}$, we can determine which item response patterns were labeled correctly or incorrectly as well as how confident $\hat{f}$ was in each decision. This approach provides a way to evaluate which observed response patterns are discrepant from the fitted IFA model.
    
    \item We can interpret the fitted NN $\hat{f}$ \parencite[e.g., using variable importance measures;][]{BreimanRF2001, Rudin2019} to determine which items were most useful for distinguishing between real and synthetic distributions.
    
    \item We can interpret the test statistic $\widehat{\acc}$ as the percentage of item response patterns that were correctly distinguished between the real and synthetic distributions.
\end{enumerate}
The values described in (a), (b), and (c) correspond to measures of person fit, piece-wise fit, and overall model fit, respectively. C2STs may therefore be viewed as uniting these different kinds of fit measures under a single framework.

\subsection{Approximate C2STs}

The C2STs described above are exact in the sense that they test the null hypothesis that the real distribution $\mathbb{P}$ and the synthetic distribution $\hat{\mathbb{P}}$ are exactly equal. In general, however, it is unlikely that any specified IFA model will exactly capture the data generating mechanism such that $\mathbb{P} = \hat{\mathbb{P}}$ \parencite[e.g.,][]{Cudeck1991, MacCallum1991}. We therefore propose a more realistic approximate C2ST (C2ST-A) for which we assume $\mathbb{P} \neq \hat{\mathbb{P}}$ and we test $H_0: \acc = 1/2 + \delta$ against $H_1: \acc > 1/2 + \delta$ where $\delta \in (0, 1/2)$ is a pre-specified value representing the degree of model error viewed as tolerable by the user. The C2ST-A is not a test of exact GOF because it does not test whether the hypothesized IFA model exactly captures the data generating mechanism. Instead, the C2ST-A is a test of approximate GOF wherein the user asserts that an IFA model that fits the data ``well enough'' should be capable of synthesizing item response patterns that can only be distinguished from real item response patterns around $100\delta \%$ of the time.

We now derive the asymptotic null distribution of $\acc$ for the C2ST-A. Since $\acc = 1/2 + \delta > 1/2$ under $H_0$, Equation (\ref{eq:3.1}) implies that $\acc_i = 1/2 + \delta_i \geq 1/2$ for $i \in I_{\test}$ where $\delta_i \in [0, 1/2]$ and ${\delta = N_{\test}^{-1} \sum_{i \in I_{\test}} \delta_i}$. In this setting, the $\acc_i$ are independent but not identically distributed Bernoulli random variables with success probabilities $p_i = 1/2 + \delta_i$. $N_{\test}\acc$ therefore follows a Poisson binomial distribution, which we follow \textcite{Ehm1991} in approximating as
\begin{equation} \label{eq:3.4}
    N_{\test}\acc \mathrel{\dot\sim} \binomial \left(N_{\test}\acc \mid N_{\test}, \bar{p} \right) \approx \mathcal{N} \big(N_{\test}\acc \mid N_{\test}\bar{p},N_{\test} \bar{p} (1 - \bar{p}) \big)
\end{equation}
for large $N_{\test}$ where $\bar{p} = N_{\test}^{-1} \sum_{i \in I_{\test}} p_i = 1/2 + N_{\test}^{-1} \sum_{i \in I_{\test}} \delta_i = 1/2 + \delta$. It follows that
\begin{equation} \label{eq:3.5}
    \acc \xrightarrow[]{d} \mathcal{N}\left(\acc \,\middle\vert\, \frac{1}{2} + \delta, \frac{\frac{1}{4} - \delta^2}{N_{\test}} \right).
\end{equation}

To derive the asymptotic alternative distribution of $\acc$, notice that under $H_1$ we have $\acc = 1/2 + \delta + \varepsilon > 1/2$ where the effect size $\varepsilon \in (0, 1/2 - \delta)$ is the magnitude of the difference between $\acc$ under $H_0$ and $\acc$ under $H_1$. Equation (\ref{eq:3.1}) implies that $\acc_i = 1/2 + \delta_i + \varepsilon_i$ for $i \in I_{\test}$ where $\delta_i \in [0, 1/2]$, $\varepsilon_i \in [0, 1/2 - \delta_i]$, ${\delta = N_{\test}^{-1} \sum_{i \in I_{\test}} \delta_i}$, and ${\varepsilon= N_{\test}^{-1} \sum_{i \in I_{\test}} \varepsilon_i}$. Then by a similar argument to the one given in the previous paragraph, we can obtain
\begin{equation} \label{eq:3.6}
    \acc \xrightarrow[]{d} \mathcal{N}\left(\acc \,\middle\vert\, \frac{1}{2} + \delta + \varepsilon, \frac{\frac{1}{4} - \delta^2 - 2 \delta \varepsilon - \varepsilon^2}{N_{\test}} \right).
\end{equation}

We now analyze the C2ST-A's power (i.e., the probability of correctly rejecting $H_0$ when $H_0$ is false) by proving the following theorem.
\begin{theorem} \label{theorem:1}
    Let $\alpha \in [0, 1]$ be the user-defined significance level (i.e., the probability of incorrectly rejecting $H_0$ when $H_0$ is true). Suppose the null and alternative distributions of $\acc$ are given by equations~\ref{eq:3.5} and~\ref{eq:3.6}, respectively. Then the power of the C2ST-A is approximately given by
    \begin{equation}
        \power(\alpha, N_{\test}, \delta, \varepsilon) \approx \Phi \left( \frac{\varepsilon \sqrt{N_{\test}} - \sqrt{\frac{1}{4} - \delta^2} \Phi^{-1}(1 - \alpha)}{\sqrt{\frac{1}{4} - \delta^2 - 2 \delta \varepsilon - \varepsilon^2}} \right). \nonumber
    \end{equation}
\end{theorem}
Proof of Theorem~\ref{theorem:1} is given in Appendix \hyperref[appendix:A]{A} and follows the approach of \textcite[][Theorem 1]{Lopez-Paz2017}.
\begin{remark}
    The approximate power of the exact C2ST was derived by \textcite[][Theorem 1]{Lopez-Paz2017} and can be alternately be derived by setting $\delta = 0$ in the power formula in Theorem~\ref{theorem:1}.
\end{remark}
We follow \textcite{Jitkrittum2016} and \textcite{Lopez-Paz2017} in setting $N_{\train} = N_{\test} = N$, which often achieves high power in practice and would achieve maximum power if $\mathbb{P}$ and $\hat{\mathbb{P}}$ differed only in means.

\subsection{C2ST-Based Relative Fit Index}

The C2STs described above measure how well the proposed IFA model reproduces the observed data (i.e., the model's absolute GOF). An alternative approach that is often used in SEM is to calculate a relative fit index (RFI) that measures the proportional improvement in fit obtained by moving to the proposed model from a more restrictive baseline model \parencite[e.g.,][]{Bentler1980a, Bentler1990, BollenFit1989, Tucker1973}. A typical baseline model posits that the observed variables are mutually independent \parencite[i.e., there are no common latent factors underlying the data; ][]{Bentler1980a} and serves as a contrasting point of reference to a model that perfectly reproduces the observed data \parencite[i.e., a saturated model; e.g.,][]{Steiger1980, Bentler1995}.

Consider a zero-factor baseline model where $P = 0$ and $\boldsymbol{\theta}_j = \boldsymbol{\alpha}_j$ for $j = 1, \ldots, J$. In this case, each respondent's response probability $\pi_{i,j,k}$ evaluated at the maximum likelihood estimate of $\boldsymbol{\theta}$ can be shown to be the observed proportion of respondents choosing response category $k$ for item $j$, which we write as $\hat{\pi}_{j, k} = N^{-1} \sum_{i = 1}^N \mathbbm{1}(x_{i, j} = k)$. We can therefore conduct a C2ST for this baseline model using $N$ synthetic samples drawn from a multinomial distribution with $K$ cells, trial size $1$, and cell probabilities $\hat{\pi}_{j,k}$. Let $\widehat{\acc}_{\prop}$ and $\widehat{\acc}_{\base}$ denote the test set accuracies for the proposed model and for the baseline model, respectively, and let $\hat{f}_{\prop}$ and $\hat{f}_{\base}$ denote the corresponding fitted NNs. We propose the following C2ST-based RFI:
\begin{equation} \label{eq:3.7}
    \rfi = 1 - \frac{M_{\prop}}{M_{\base}} \cdot  \frac{\Delta_{\prop}}{\Delta_{\base}},
\end{equation}
where $\Delta_{\prop} = \widehat{\acc}_{\prop} - 1/2$, $\Delta_{\base} = \widehat{\acc}_{\base} - 1/2$, $M_{\prop}$ is the number of fitted parameters in the proposed model, and $M_{\base}$ is the number of fitted parameters in the baseline model. The ratio $\Delta_{\prop} / \Delta_{\base}$ is a measure of the proportional change in misfit obtained by moving to the proposed model from the baseline model, while the ratio $M_{\prop} / M_{\base}$ is a penalty that increases with the number of fitted parameters in the proposed model (i.e., it rewards parsimony in the proposed model).

We motivate our definition of $\rfi$ by considering the fit index's behavior in the typical setting where the baseline model obtains less-than-perfect fit (i.e., when $\widehat{\acc}_{\base} > 1/2$). In the common scenario that the proposed model fits the same or better than the baseline model (i.e., when $\widehat{\acc}_{\prop}$ varies between $\widehat{\acc}_{\base}$ and $1/2$), $\rfi$ varies between $1 - M_{\prop} / M_{\base}$ and one with values closer to one indicating better fit. $\rfi$ is much larger than one when $\widehat{\acc}_{\prop}$ is much smaller than $1/2$, which may occur when proposed model fits the data well but $\hat{f}_{\prop}$ has overfitted the training data. In the infrequent event that the proposed model fits worse than the baseline model (i.e., when $\widehat{\acc}_{\prop} > \widehat{\acc}_{\base}$), $\rfi$ is smaller than $1 - M_{\prop} / M_{\base}$. $\rfi$ therefore behaves somewhat analogously to other nonnormed fit indices such as the Tucker-Lewis Index \parencite{Tucker1973} that distinguish between less-than-perfect model fit, perfect model fit, and overfitting by being smaller than one in the first case, close to one in the second case, and much larger than one in the third case.

We next consider the effect of sample size on $\rfi$. \textcite{BollenFit1989} notes that GOF indices may be influenced by sample size either (a) when $N$ enters the calculation of the index or (b) when the mean of the sampling distribution of the index is related to $N$. For (a), notice that although both $\widehat{\acc}_{\prop}$ and $\widehat{\acc}_{\base}$ include factors of $N_{\test}^{-1}$,  these factors cancel in the ratio $\Delta_{\prop} / \Delta_{\base}$, indicating that (a) does not hold. For (b), notice that when $N$ is very small, both $\hat{f}_{\prop}$ and $\hat{f}_{\base}$ may fail to capture any relationships in the training data. In this case, both $\widehat{\acc}_{\prop}$ and $\widehat{\acc}_{\base}$ will be close to $1/2$ and $\rfi$ will be close to $1 - M_{\prop} / M_{\base}$. As $N$ increases, $\hat{f}_{\prop}$ and $\hat{f}_{\base}$ should capture any relationships in the training data and $\widehat{\acc}_{\prop}$ and $\widehat{\acc}_{\base}$ should come close to their population values, suggesting that (b) holds. We provide empirical evidence that the mean of the sampling distribution of $\rfi$ depends on $N$ in Sect. \ref{EvalIFASim}. Specifically, we observe that C2ST-RFI gives a less optimistic assessment of fit as $N$ increases, which is analogous to the approximate C2ST's power increasing with $N$.

We also consider the probability limit of the proposed fit index. Assume that $\plim_{N \rightarrow \infty} (\acc_{\prop}) = 1/2 + \delta$ and $\plim_{N \rightarrow \infty} (\acc_{\base}) = 1/2 + \delta + \varepsilon$ where $\delta \in [0, 1/2]$ and $\varepsilon \in [0, 1/2 - \delta]$. This assumption is mild and states that the accuracies obtained by $\hat{f}_{\prop}$ and $\hat{f}_{\base}$ should come close to specific constants as $N$ grows large, with $\hat{f}_{\prop}$ obtaining the same or better accuracy than $\hat{f}_{\base}$.\footnote{A similar assumption that $\plim_{N,\; J \rightarrow \infty} (\acc) = 1/2 + \delta$ for some $\delta > 0$ is used by \textcite{Kim2021} to prove that C2STs are consistent (i.e., have power approaching one) in the high-dimensional setting where both $N$ and $J$ tend to infinity.} Then $\plim_{N \rightarrow \infty} (\rfi) = 1 - \lbrack M_{\prop} / M_{\base} \rbrack \cdot \lbrack \delta / (\delta + \varepsilon) \rbrack = c$ where $c$ is a constant which equals one when the proposed model is correct (i.e., when $\delta = 0)$ and is smaller than one otherwise.

Finally, we comment on how to interpret numerical values of C2ST-RFIs. In practice, cutoff values are used to interpret fit indices and thereby evaluate model fit. The development of adequate ``rules of thumb'' cutoff criteria that cover a wide range of realistic data and model conditions requires extensive empirical study \parencite[e.g.,][]{Hu1999} and is beyond the scope of this work. Based on limited initial experimentation, however, we consider a provisory cutoff of $\rfi > 0.9$ to indicate good fit. We empirically investigate the performance of this cutoff in Sect. \ref{EvalIFASim}.

\section{Implementation and Experiments}

\subsection{Implementation}

I-WAVE is implemented using the Python package DeepIRTools \parencite[Version 0.2.1;][]{UrbanDeepIRTools}. Although DeepIRTools supports GPU computing to accelerate fitting, we follow \textcite{UrbanBauer2021} in opting for CPU computing to enable fairer comparisons with other methods and to assess performance using hardware that is more typically available to social scientists. Experiments are conducted on a computer with a 2.8 GHz Intel Core i7 CPU and 16 GB of RAM. Code to reproduce all experiments is readily available at \url{https://github.com/cjurban/DeepConfirmatoryIFA}.

We now provide hyperparameter settings for I-WAVE. For heuristic justifications for most settings, see \textcite[][Sect. 5]{UrbanBauer2021}. Optimization hyperparameters include the AMSGrad learning rate and minibatch size, which we set to $5 \times 10^{-3}$ and $128$, respectively. The NN inference model has a single hidden layer of size $100$ and exponential linear unit (ELU) activation functions \parencite{Clevert2016}. We follow \textcite{UrbanBauer2021} in determining convergence by calculating the average $\iwelbo$ every $100$ fitting iterations and stopping fitting if this average value does not decrease after $100$ such calculations. The NN inference model is initialized such that the posterior means $\boldsymbol{\mu}_i$ and variances $\boldsymbol{\sigma}_i^2$ are close to $\mathbf{0}$ and $\mathbf{1}$, respectively. Unconstrained loadings vectors $\boldsymbol{\beta}_j'$ are initialized close to $\mathbf{1}$, while intercepts vectors $\boldsymbol{\alpha}_j$ are initialized such that the cumulative standard logistic distribution between consecutive elements is fixed. An additional detail not discussed by \textcite{UrbanBauer2021} that must be addressed in the confirmatory setting is initializing the hyperspherical parameterization $\boldsymbol{\Theta}$ of the factor correlation matrix $\boldsymbol{\Sigma}$. To ensure that $\boldsymbol{\Sigma}$ is well-conditioned at the start of fitting, we choose $\boldsymbol{\Theta}$ such that $\boldsymbol{\Sigma} = \mathbf{I}_P$ by setting $\vartheta_{p, p'} = \pi /2$ for $p = 1, \ldots, P$, $p' = 1, \ldots, p - 1$.

C2STs are programmed using the machine learning library PyTorch \parencite[Version 1.10.1][]{Paszke2017}. NN classifiers have ELU activation functions and are initialized such that predicted probabilities are close to $1/2$. We fit NNs using the same optimization hyperparameters and convergence criterion used for I-WAVE. To mitigate possible overfitting, we (1) use NNs with a single hidden layer of size $20$ and (2) stop fitting if the default convergence criterion is not obtained within $\lfloor \num{100000} \cdot 128 / N_{\train} \rfloor$ stochastic gradient steps. Since NNs are not directly interpretable, we use a permutation importance (PI) method to interpret which items are most useful for distinguishing between real and synthetic distributions \parencite{BreimanRF2001, Fisher2019}. Let $\mathbf{U}$ be an $N \times J$ matrix whose $i^{\mathrm{th}}$ row is $\mathbf{u}^\top_i$ for $i \in I_{\test}$. For each item $j = 1, \ldots, J$ and each repetition $t =1, \ldots, T$, shuffle the $j^{\mathrm{th}}$ column of $\mathbf{U}$ to generate a corrupted matrix $\widetilde{\mathbf{U}}_{j, t}$ whose $i^{\mathrm{th}}$ row is $\widetilde{\mathbf{u}}^\top_{i, j, t}$. The PI for item $j$ is defined as the mean decrease in test set accuracy for $\hat{f}$ when the responses to item $j$ are shuffled:
\begin{equation} \label{eq:4.2}
    \imp_j = \widehat{\acc} - \frac{1}{T} \sum_{t = 1}^T \frac{1}{N_{\test}} \sum_{i \in I_{\test}} \mathbbm{1} \bigg( \mathbbm{1} \bigg( \hat{f}(\widetilde{\mathbf{u}}_{i, j, t}) > \frac{1}{2} \bigg) = l_i \bigg) = \widehat{\acc} - \frac{1}{T} \sum_{t = 1}^T \widetilde{\acc}_{j, t}.
\end{equation}
Since the shuffling procedure breaks the relationship between item $j$ and the class label, $\imp_j$ measures how much $\hat{f}$ depends on item $j$. PI may be viewed as a piece-wise fit assessment method wherein each $\imp_j$ measures how well the proposed IFA model fits item $j$, with larger $\imp_j$ indicating worse fit. We set the number of repetitions to $T = 10$ for all experiments.

\subsection{Empirical Example}

We demonstrate the proposed methods' computational efficiency in the large-scale setting and obtain data generating parameters for simulation studies by analyzing $\num{1015342}$ responses to the $50$-item International Personality Item Pool five-factor model \parencite[IPIP-FFM;][]{Goldberg1999}, the same data considered by \textcite{UrbanBauer2021} for exploratory IFA using I-WAVE. The IPIP-FFM is designed to measure the Big Five personality factors of openness, conscientiousness, extraversion, agreeableness, and emotional stability, making this data well suited to a more confirmatory approach. Each factor is measured by $10$ five-category items anchored by “Disagree” (1), “Neutral” (3), and “Agree” (5). The data were downloaded from the Open-Source Psychometrics Project (\url{https://openpsychometrics.org/}) and pre-processed similarly to \textcite[][Sect. 6.1]{UrbanBauer2021}, resulting in an analytic sample containing $N = \num{548389}$ responses. Reverse worded items were recoded so that the highest numerical response category indicated a high level of the corresponding factor.

\subsubsection{A Five-Factor Model}

We used I-WAVE to fit a five-factor confirmatory IFA model with correlated factors to the IPIP-FFM data. We set the number of IW samples to $R = 10$ based on \citeauthor{UrbanBauer2021}'s (\citeyear{UrbanBauer2021}) finding that a moderate number of samples performs well in practice. We fitted the data set $10$ times to investigate whether parameter estimation and GOF assessment results were stable across random starts.

Factor loadings and correlations from the fitted model that attained the highest IW-ELBO across random starts --- henceforth called the reference model --- are reported in Tables~\ref{tab:1} and ~\ref{tab:2}, respectively. All loadings were positive, which fit with the confirmatory design of the measurement scale. Factor correlations aligned with the typical finding that emotional stability is negatively correlated with the other factors. The mean fitting time was $167$ seconds ($SD = 92$ seconds), which is quite fast given the large sample size. Relative to the reference model, mean loadings root-mean-square error (RMSE) was $0.03$ ($SD = 0.01$), mean intercepts RMSE was $0.03$ ($SD = 0.01$), and mean factor correlation RMSE was $0.02$ ($SD = 0.01$), suggesting that fitting was stable.\footnote{We treat the model attaining the highest IW-ELBO as ground truth in line with how an optimal solution is often selected from multiple random starts for mixture models \parencite[e.g.,][]{Biernacki2003}. Were we to alternatively treat the mean parameter estimates across random starts as ground truth, the corresponding variability estimates would be smaller than those reported here, although this approach would downplay the impact of possible local maxima on parameter estimate stability.}

\begin{table}
    \centering
    \caption{Factor Loadings for IPIP-FFM Data Set\label{tab:1}}
    \begin{tabular}{lccccccccccl}
        \toprule
         Factor & \multicolumn{10}{c}{Loading} & \multicolumn{1}{c}{Items} \\
        \midrule
        \multicolumn{12}{c}{\textit{Five-Factor Model}} \\
        Extraversion & 1.90 & 2.03 & 1.95 & 2.34 & 2.41 & 1.50 & 2.28 & 1.43 & 1.57 & 2.16 & 1--10 \\
        Emotional Stability & 1.95 & 1.30 & 1.51 & $\phantom{2}$.97 & 1.24 & 2.28 & 2.25 & 2.55 & 1.90 & 1.76 & 11--20 \\
        Agreeableness & 1.38 & 1.63 & $\phantom{2}$.73 & 2.77 & 2.05 & 1.43 & 2.03 & 1.50 & 2.13 & 1.05 & 21--30 \\
        Conscientiousness & 1.57 & 1.42 & $\phantom{2}$.79 & 1.70 & 1.66 & 1.78 & 1.23 & 1.28 & 1.47 & $\phantom{2}$.93 & 31--40 \\
        Openness & 1.18 & 1.42 & 1.42 & 1.24 & 1.72 & 1.52 & 1.07 & 1.04 & $\phantom{2}$.74 & 2.30 & 41--50 \\
        \multicolumn{12}{c}{\textit{Seven-Factor Model}} \\
        Extraversion & 1.87 & 2.08 & 1.91 & 2.34 & 2.42 & 1.49 & 2.26 & 1.43 & 1.56 & 2.21 & 1--10 \\
        Emotional Stability & 2.27 & 1.46 & 1.76 & $\phantom{2}$.99 & 1.24 & 2.35 & 2.85 & 3.23 & 1.82 & 1.75 & 11--20 \\
        Agreeableness & 1.42 & 1.61 & $\phantom{2}$.73 & 2.78 & 2.03 & 1.39 & 2.01 & 1.52 & 2.199 & 1.03 & 21--30 \\
        Conscientiousness & 1.60 & 1.43 & $\phantom{2}$.82 & 1.66 & 1.65 & 1.77 & 1.24 & 1.27 & 1.43 & $\phantom{2}$.96 & 31--40 \\
        Openness & 1.56 & 1.36 & 1.54 & 1.25 & 1.77 & 1.68 & $\phantom{2}$.96 & 1.34 & $\phantom{2}$.77 & 2.58 & 41--50 \\
        Doublet 1 & & & & & & & 2.47 & 2.47 & & & 11--20 \\
        Doublet 2 & 2.41 & & & & & & & 2.41 & & & 41--50 \\
        \bottomrule
    \end{tabular}
\end{table}

\begin{table}
    \centering
    \addtolength{\tabcolsep}{3pt}
    \caption{Factor Correlations for IPIP-FFM Data Set} \label{tab:2}
        \begin{tabular}{lcccccccc}
            \toprule
                 & \multicolumn{7}{c}{Factor} \\ \cmidrule{2-8}
            Factor & {$1$} & {$2$} & {$3$} & {$4$} & {$5$} & {$6$} & {$7$} \\
            \midrule
            \multicolumn{8}{c}{\textit{Five-Factor Model}} \\
            1. Extraversion & $\phantom{-}$1.00 & & & & & & \\
            2. Emotional Stability & $\phantom{1}-$.25 & $\phantom{-}$1.00 & & & & & \\
            3. Agreeableness & $\phantom{1-}$.39 & $\phantom{1}-$.03 & $\phantom{-}$1.00 & & & & \\
            4. Conscientiousness & $\phantom{1-}$.07 & $\phantom{1}-$.30 &$\phantom{1-}$.14 & $\phantom{-}$1.00 & & & \\
            5. Openness & $\phantom{1-}$.22 & $\phantom{1}-$.12 &$\phantom{1-}$.16 &$\phantom{1-}$.05 & $\phantom{-}$1.00 & & \\
            \multicolumn{8}{c}{\textit{Seven-Factor Model}} \\
            1. Extraversion & $\phantom{-}$1.00 & & & & & & \\
            2. Emotional Stability & $\phantom{1}-$.27 & $\phantom{-}$1.00 & & & & & \\
            3. Agreeableness & $\phantom{1-}$.35 &$\phantom{1-}$.00 & $\phantom{-}$1.00 & & & & \\
            4. Conscientiousness & $\phantom{1-}$.11 & $\phantom{1}-$.27 &$\phantom{1-}$.15 & $\phantom{-}$1.00 & & & \\
            5. Openness & $\phantom{1-}$.21 & $\phantom{1}-$.12 &$\phantom{1-}$.16 &$\phantom{1-}$.09 & $\phantom{-}$1.00 & & \\
            6. Doublet 1 & $\phantom{1-}$.00 &$\phantom{1-}$.00 &$\phantom{1-}$.00 &$\phantom{1-}$.00 &$\phantom{1-}$.00 & $\phantom{-}$1.00 & \\
            7. Doublet 2 & $\phantom{1-}$.00 &$\phantom{1-}$.00 &$\phantom{1-}$.00 &$\phantom{1-}$.00 &$\phantom{1-}$.00 &$\phantom{1-}$.00 & $\phantom{-}$1.00 \\
            \midrule[\heavyrulewidth]
        \end{tabular}
\end{table}

We assessed overall fit for the proposed five-factor model (FFM) across random starts using exact C2STs and C2ST-As. For C2ST-As, we set $\delta = 0.05$ to test $H_0 : \acc = 0.55$ against $H_1 : \acc > 0.55$, which amounts to testing whether the proposed FFM is capable of synthesizing item response patterns that can only be distinguished from real item response patterns $55 \%$ of the time (i.e., at slightly better than chance). C2ST results are presented in Table~\ref{tab:3}. Fitted NNs obtained high test set accuracies ($M = 0.81$, $SD = 0.01$) and all C2STs rejected $H_0$ at signficance level $\alpha = 0.05$ ($\hat{p} < 0.0001$), suggesting that the FFM did not fit the data ``well enough''. Importantly, tests were fast: accounting for both sampling synthetic data and fitting the NN, C2STs took around one minute.

\begin{table}
    \centering
    \addtolength{\tabcolsep}{3pt}
    \caption{Classifier Two-Sample Test Results for IPIP-FFM Data Set} \label{tab:3}
    \begin{threeparttable}
        \begin{tabular}{lc@{\hspace{0.15cm}}ccc@{\hspace{0.1cm}}ccc@{\hspace{0.1cm}}cc}
            \toprule
            & & \multicolumn{2}{c}{Baseline Model} & & \multicolumn{2}{c}{Five-Factor Model} & & \multicolumn{2}{c}{Seven-Factor Model} \\ \cmidrule{3-4} \cmidrule{6-7} \cmidrule{9-10}
            & & {$M$} & {$SD$} & & {$M$} & {$SD$} & & {$M$} & {$SD$} \\
            \midrule
            Sampling Time & & $\phantom{11}$.8$\phantom{1}$ & {$\phantom{11<\;}$}.2 & & $\phantom{1}$6.0$\phantom{1}$ & $\phantom{1}$2.0$\phantom{1}$ & & $\phantom{1}$5.5$\phantom{1}$ & $\phantom{1}$1.0$\phantom{1}$ \\
            Fitting Time & & 56$\phantom{.11}$ & {$\phantom{<\;}$}26$\phantom{.11}$ & & 48$\phantom{11}$ & 19$\phantom{.11}$ & & 58$\phantom{.11}$ & 21$\phantom{.11}$ \\
            Test Set Accuracy & & $\phantom{11}$.93 & $\phantom{11}<\;$.01 & & $\phantom{11}$.81 & $\phantom{11}$.01 & & $\phantom{11}$.80 & $\phantom{11}$.01 \\
            C2ST-RFI & & & & & $\phantom{11}$.25 & $\phantom{11}$.02 & & $\phantom{11}$.28 & $\phantom{11}$.02 \\
            \midrule[\heavyrulewidth]
        \end{tabular}
        \begin{tablenotes}
            \item \textit{Note.} ``Sampling Time'' refers to time required to sample synthetic data from the model. All times are given in seconds.
        \end{tablenotes}
    \end{threeparttable}
\end{table}

\begin{figure}
    \centering
    \begin{subfigure}{0.8\textwidth}
        \includegraphics[width=\textwidth]{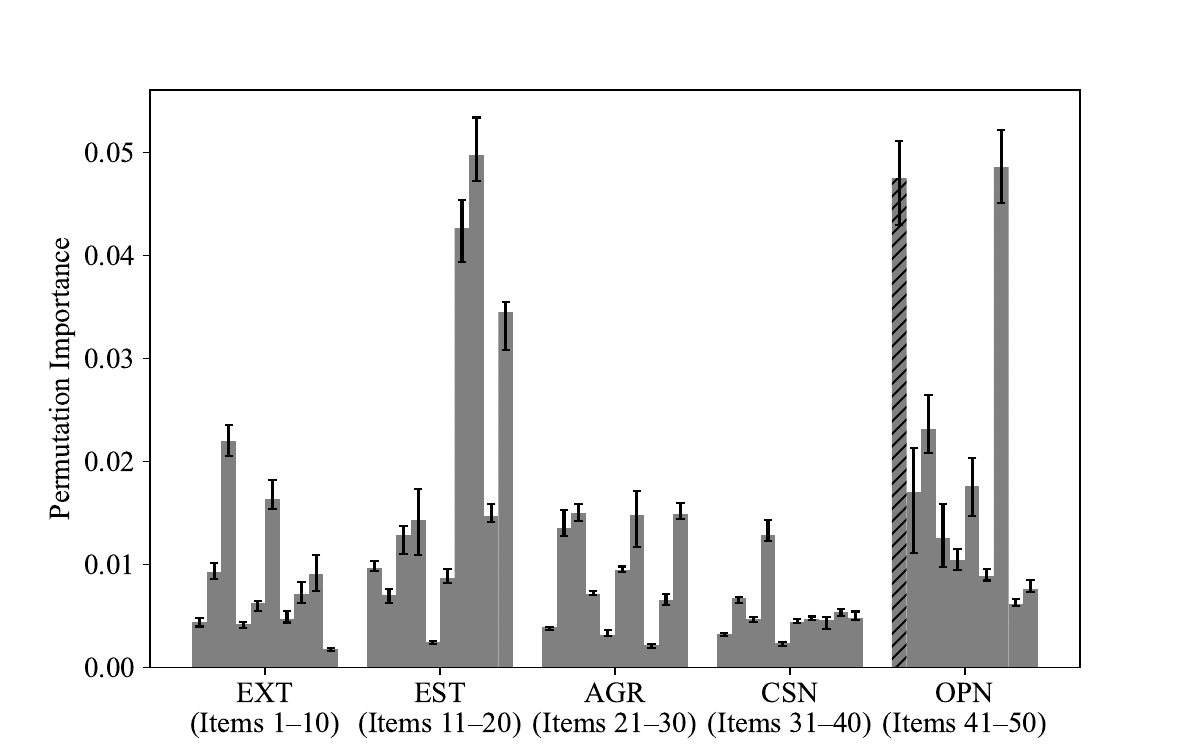}
        \caption{Five-factor model.} \label{fig:2a}
    \end{subfigure}
    
    \begin{subfigure}{0.8\textwidth}
        \includegraphics[width=\textwidth]{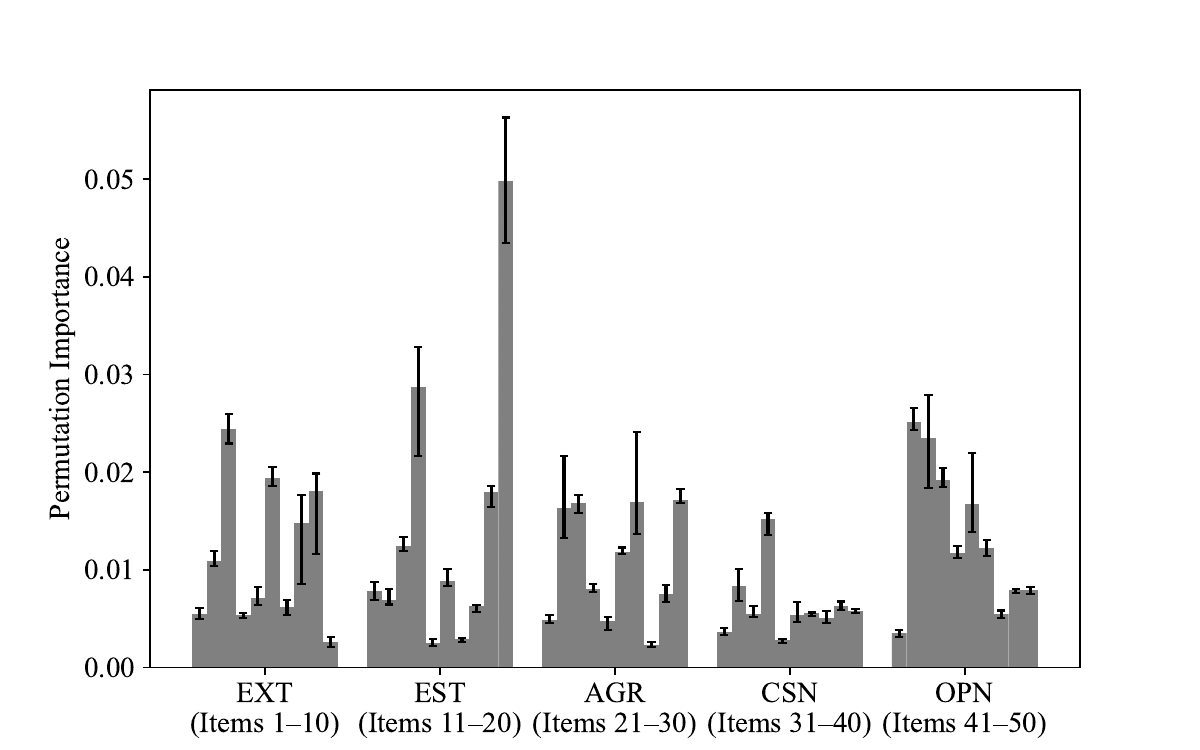}
        \caption{Seven-factor model.} \label{fig:2b}
    \end{subfigure}
    \caption{Permutation importances for each IPIP-FFM item across $10$ random starts. Items are grouped by their corresponding factor. Bar heights indicate medians, while error bars indicate $25\%$ and $75\%$ quantiles. Hatched bars indicate the five poorest fitting items for the five-factor model. EXT = extraversion, EST = emotional stability, AGR = agreeableness, CON = conscientiousness, OPN = openness.} \label{fig:2}
\end{figure}

We next assessed piece-wise fit by computing PIs for fitted classifiers accross random starts, which are displayed in Figure~\ref{fig:2a}. PIs show that NN classifiers relied relatively heavily on five specific items --- items $17$, $18$, and $20$, which measure emotional stability, and items $41$ and $48$, which measure openness. This finding suggests that the FFM's overall lack of fit is most heavily influenced by its piece-wise lack of fit to items $17$, $18$, $20$, $41$, and $48$.

Although the above results indicate that the FFM did not fit the data well in an absolute sense, similar findings might be expected for nearly any \textit{a priori} model. It would therefore be useful to assess the FFM's value in a relative sense by contrasting it with a baseline model. To this end, we computed ${\rfi}\mathrm{s}$ for the fitted NNs across random starts. As shown in Table~\ref{tab:3}, NNs obtained high baseline model test set accuracies ($M = 0.93$, $SD < 0.01$), suggesting that the FFM fit the data better than the zero-factor baseline model. This finding was reflected in the ${\rfi}\mathrm{s}$, which fell between $1 - M_{\prop} / M_{\base} = -0.05$ and one ($M = 0.25$, $SD = 0.02$). No C2ST-RFIs exceeded the provisory cutoff of $0.9$ suggested as an indicator of good fit. The additional RFI computations were also fast, taking close to an additional minute. 

\subsubsection{A Seven-Factor Model}

We investigated the five poorest fitting items flagged by PIs with the goal of improving model fit. As shown in Table~\ref{tab:4}, wordings for item pair $17$ and $18$ as well as for item pair $41$ and $48$ are similar. To account for possible local dependence between these similarly worded pairs, we modeled each pair using an additional orthogonal ``doublet'' factor that was only measured by its corresponding two items and whose loadings were constrained to be equal to ensure model identification. The resulting seven-factor model (SFM) was fitted with $10$ random starts using the same hyperparameters as the FFM.

\begin{table}
    \addtolength{\tabcolsep}{3pt}
    \caption{Wordings for Five Poorest Fitting IPIP-FFM Items} \label{tab:4}
        \begin{tabular}{ll}
            \toprule
            Item & \multicolumn{1}{c}{Wording} \\
            \midrule
            17 & I change my mood a lot. \\
            18 & I have frequent mood swings. \\
            20 & I often feel blue. \\
            41 & I have a rich vocabulary. \\
            48 & I use difficult words. \\
            \midrule[\heavyrulewidth]
        \end{tabular}
\end{table}

The SFM loadings and factor correlation estimates are given in Tables~\ref{tab:1} and~\ref{tab:2}, respectively, and are largely similar to the FFM estimates for the non-doublet factors. Fitting remained fast ($M = 182$ seconds, $SD = 40$ seconds) and stable (relative to the seven-factor reference model, loadings RMSE $M = 0.03$, $SD = 0.01$; intercepts RMSE $M = 0.03$, $SD = 0.01$; and factor correlation RMSE $M = 0.01$, $SD = 0.01$).

C2ST results for the SFM in Table~\ref{tab:3} suggest that overall fit improved marginally relative to the FFM. In particular, mean test set accuracy was slightly lower for the SFM, suggesting that the SFM fit the data slightly better. However, all C2STs again rejected $H_0$ at $\alpha = 0.05$ ($\hat{p} < 0.0001$), suggesting that the SFM also failed to fit the data ``well enough''. The SFM's fit relative to the zero-factor baseline also improved only marginally: mean C2ST-RFIs were slighly higher for the SFM than for the FFM, although these values remained far from the provisory cutoff of $0.9$. Computation for all tests and fit indices remained fast.

Although the SFM appeared to improve overall fit only marginally relative to the FFM, piece-wise fit appeared to improve more substantially. This improvement is evident in the SFM's PIs, which are shown in Figure~\ref{fig:2b}. In particular, PIs for the flagged item pairs are drastically lower for the SFM than for the FFM.

\subsection{Simulation Studies}

\subsubsection{Evaluating I-WAVE}

We investigate confirmatory I-WAVE's parameter recovery and computational efficiency as the number of IW samples $R$ increases and the log-likelihood approximation improves. The data generating model has $P = 5$ factors and $J = 50$ $5$-category items. Generating parameters are rounded estimates from the five-factor reference model in the empirical example. We consider $R = 1$, $10$, and $100$ as well as $N = 500$, $\num{2500}$, $\num{12500}$, and $\num{62500}$, resulting in $12$ total simulation settings for each $R$ and $N$ combination. We conduct $100$ replications at each setting. All analyses reused the optimization and inference model hyperparameters from the empirical example.

\begin{figure}
    \centering
    \begin{subfigure}{0.5\textwidth}
    \includegraphics[width=\textwidth]{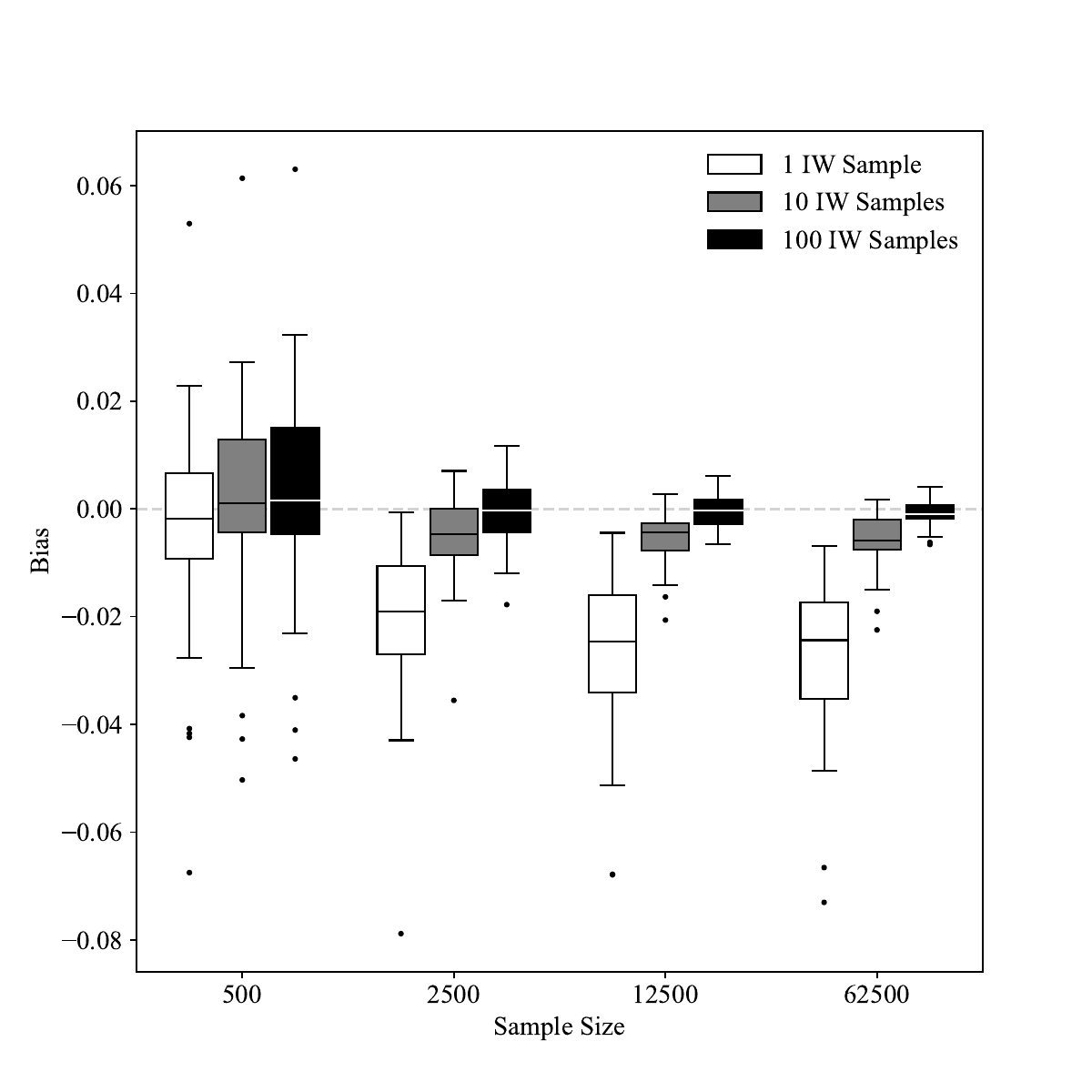}
        \caption{Factor loadings.} \label{fig:3a}
    \end{subfigure}%
    \hfill
    \begin{subfigure}{0.5\textwidth}
        \includegraphics[width=\textwidth]{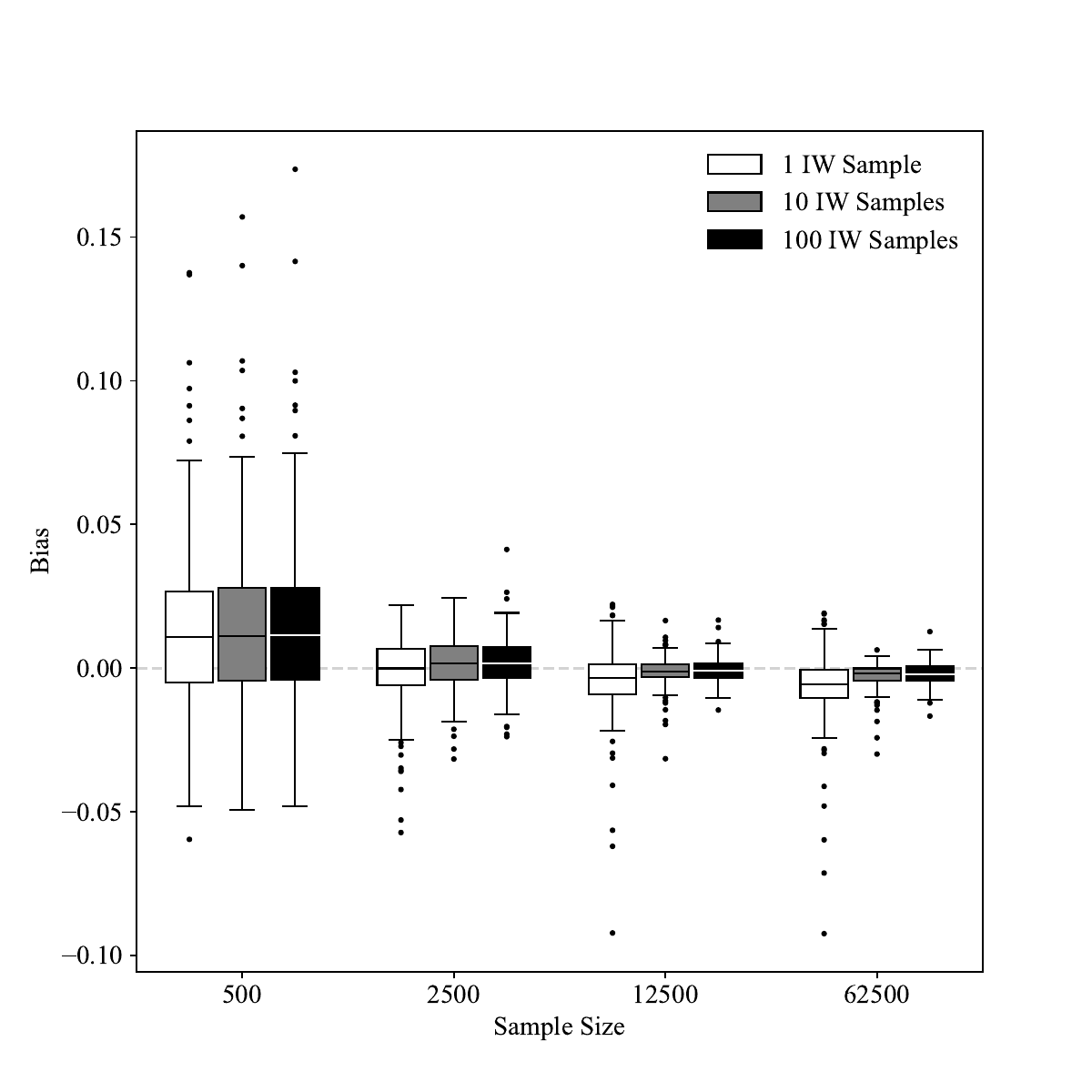}
        \caption{Category intercepts.} \label{fig:3b}
    \end{subfigure}

    \begin{subfigure}{0.5\textwidth}
        \includegraphics[width=\textwidth]{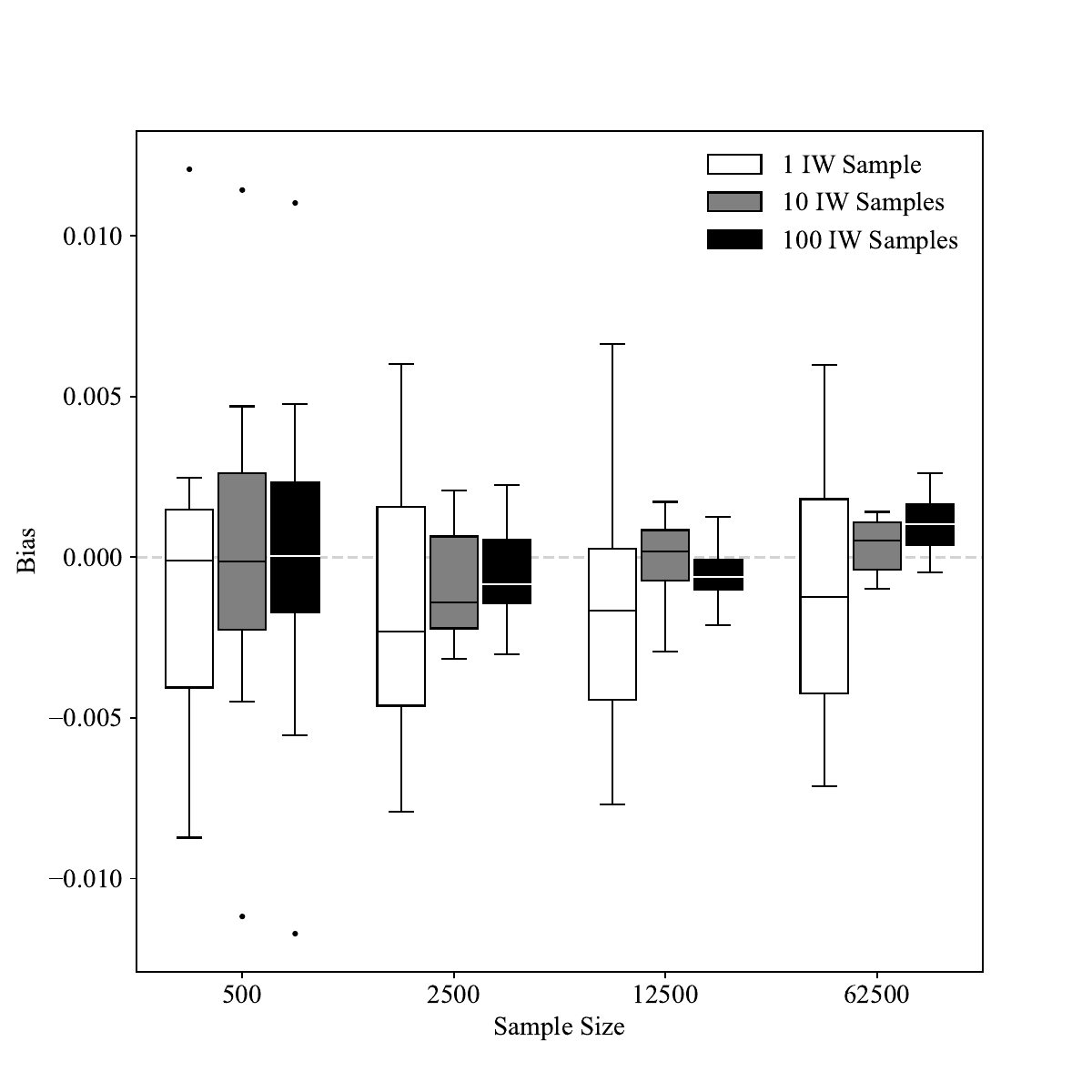}
        \caption{Factor correlations.} \label{fig:3c}
    \end{subfigure}
    \caption{Parameter bias for the importance-weighted amortized variational estimator (I-WAVE). Three settings for the number of importance-weighted (IW) samples are compared.} \label{fig:3}
\end{figure}

\begin{figure}
    \centering
    \begin{subfigure}{0.5\textwidth}
    \includegraphics[width=\textwidth]{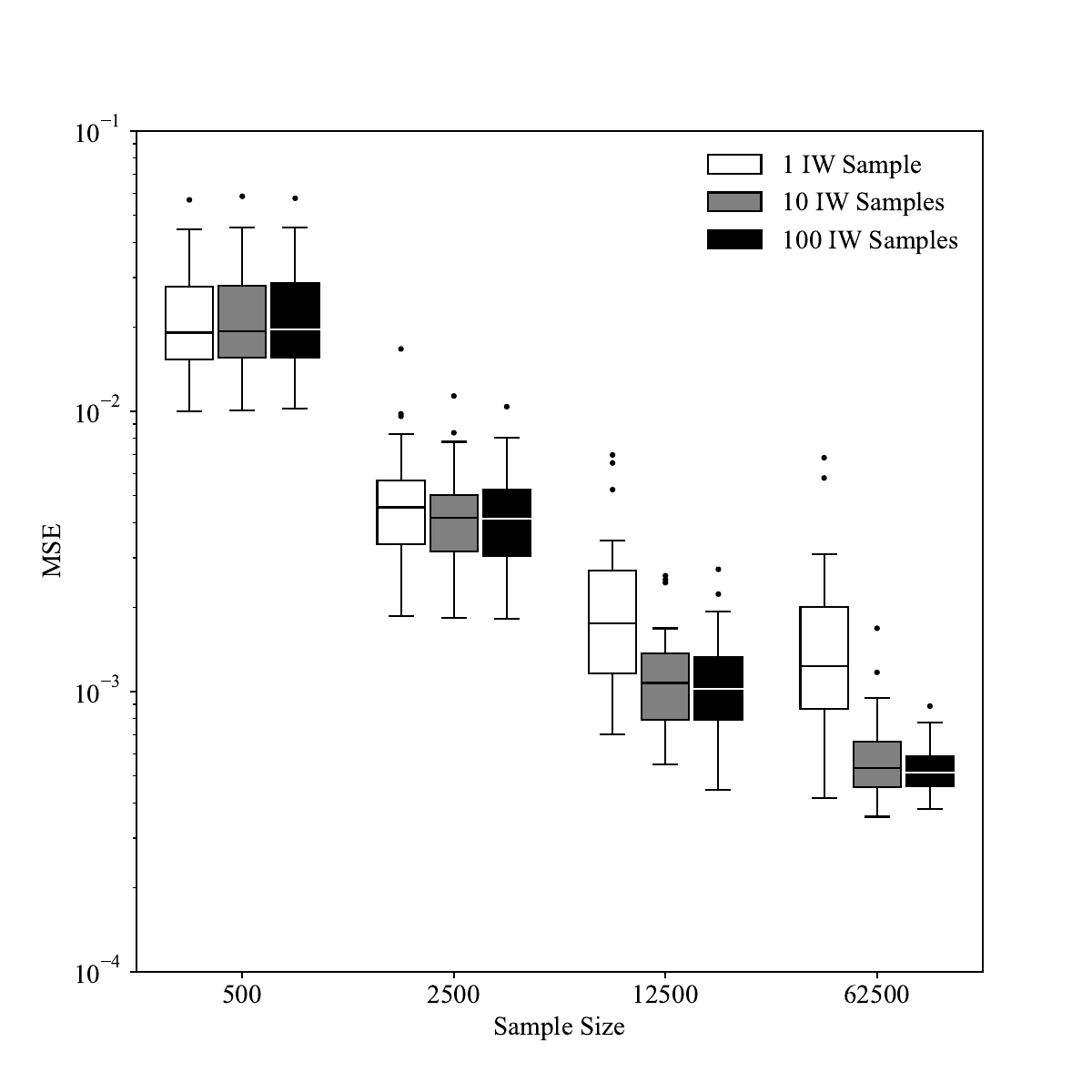}
        \caption{Factor loadings.} \label{fig:4a}
    \end{subfigure}%
    \hfill
    \begin{subfigure}{0.5\textwidth}
        \includegraphics[width=\textwidth]{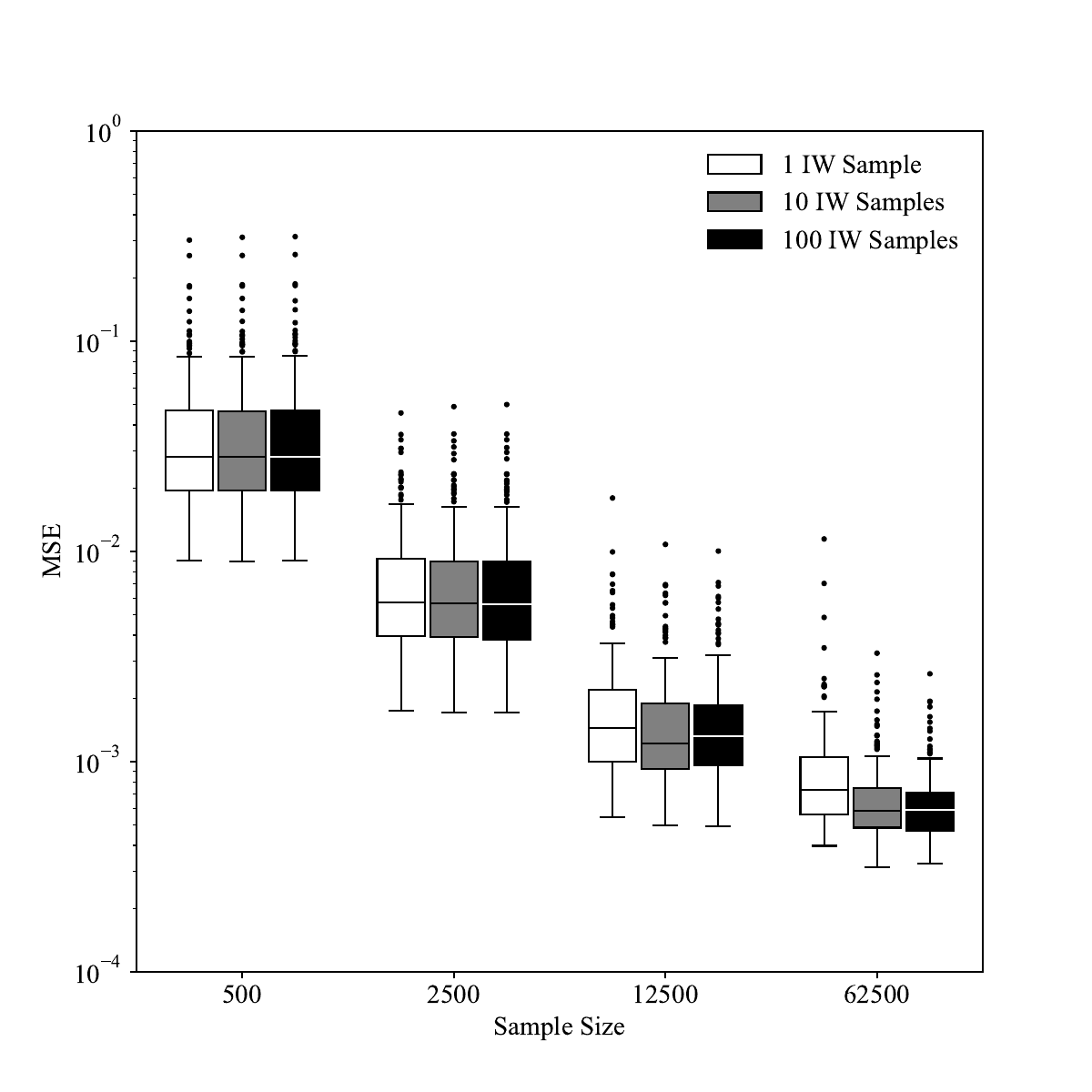}
        \caption{Category intercepts.} \label{fig:4b}
    \end{subfigure}

    \begin{subfigure}{0.5\textwidth}
        \includegraphics[width=\textwidth]{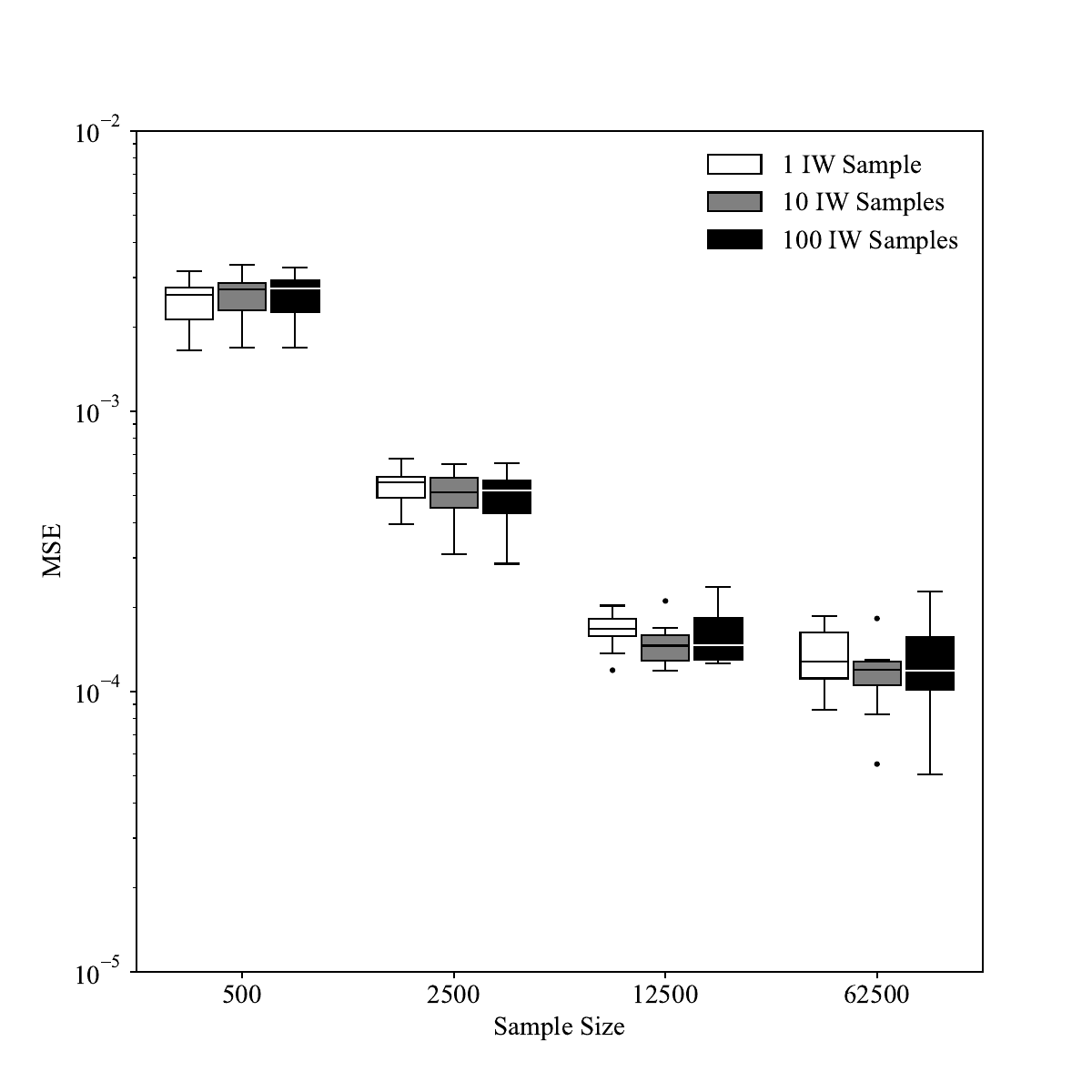}
        \caption{Factor correlations.} \label{fig:4c}
    \end{subfigure}
    \caption{Parameter mean squared error (MSE) for I-WAVE.} \label{fig:4}
\end{figure}

Parameter recovery was assessed by computing the bias for each parameter as the mean deviation of the estimated parameter from the data generating parameter across replications, that is, $\bias(\hat{\xi}, \xi) = 100^{-1} \sum_{a = 1}^{100} [ \hat{\xi}^{(a)} - \xi ]$ where $\hat{\xi}^{(a)}$ is the estimated parameter at replication $a$ and $\xi$ is the data generating parameter. We also computed mean squared error (MSE) for each parameter as $\mse(\hat{\xi}, \xi) = 100^{-1} \sum_{a = 1}^{100} [ \hat{\xi}^{(a)} - \xi ]^2$. Boxplots of parameter biases and MSEs for each simulation setting are displayed in Figures~\ref{fig:3} and~\ref{fig:4}, respectively, with separate plots for factor loadings, factor correlations, and intercepts. All estimates become more accurate as the sample size increases. Estimates are somewhat biased for $R = 1$ but appear relatively unbiased for $R \geq 10$. MSE tends to decrease with increasing $R$ for each $N$ setting. We note that a single replication in the $(R, N) = (1, 500)$ setting diverged and was excluded from these analyses.

\begin{figure}
	\centering
	\includegraphics[width=0.6\textwidth]{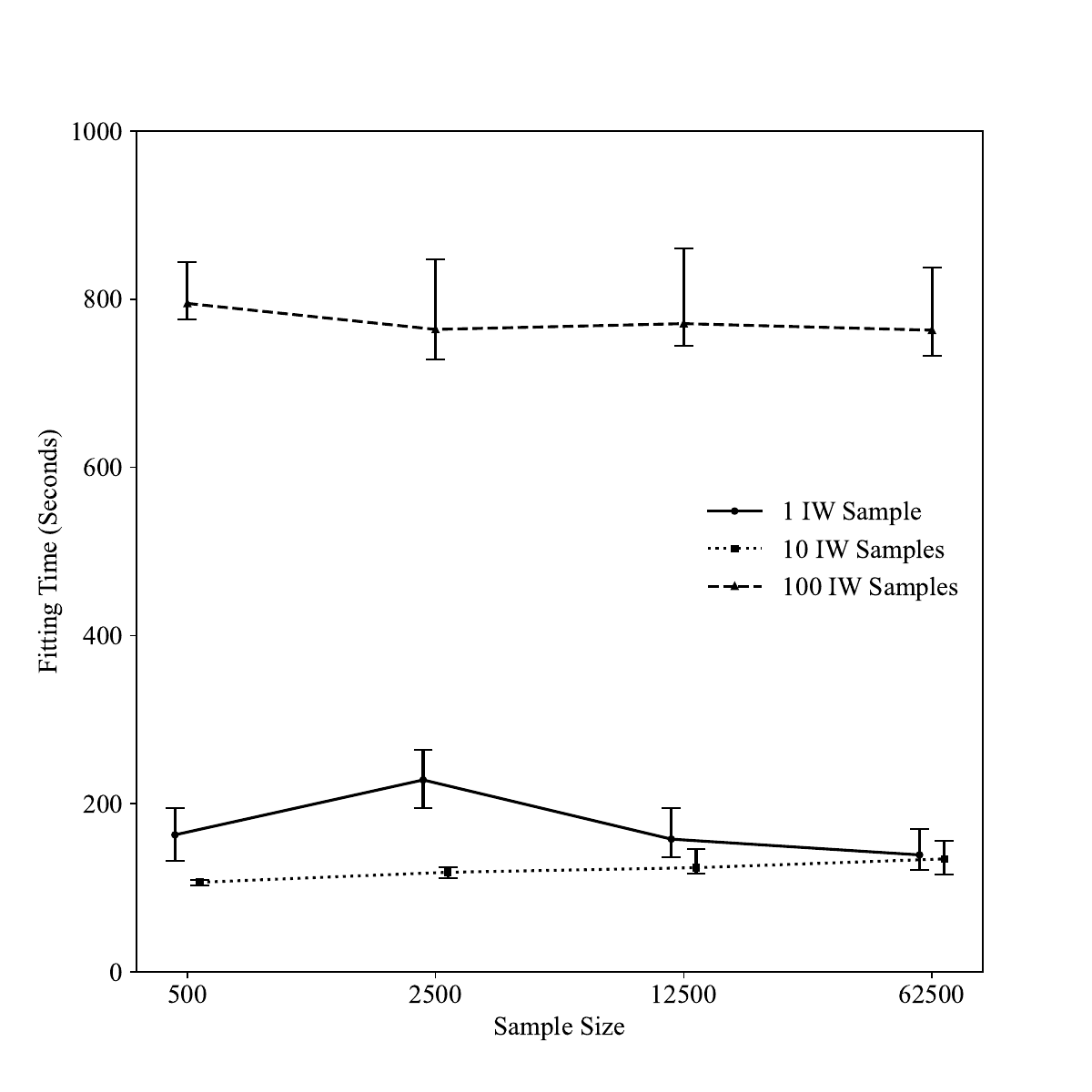}
	\caption{Fitting times for I-WAVE. Markers indicate medians, while error bars indicate $25\%$ and $75\%$ quantiles.} \label{fig:5}
\end{figure}

Line plots of fitting times for each simulation setting are displayed in Figure~\ref{fig:5}. Median fitting time decreases from around three minutes to around two minutes as $R$ increases from $1$ to $10$, then increases to around $13$ minutes as $R$ increases from $10$ to $100$. For fixed $R$, median fitting time remains close to constant as $N$ increases. These results demonstrate that I-WAVE is computationally efficient even when the sample size is very large.

\subsubsection{Comparing I-WAVE to MH-RM}

In this study, we compare I-WAVE to the MML estimator implemented via MH-RM in a setting where the number of factors is large. We use the MH-RM implementation from the R package mirt, which has core functions written in both R and C++ \parencite[Version 1.32.1][]{Chalmers2012}. The data generating model has $P = 10$ factors measured by $J = 100$ $5$-category items. Generating parameters are again rounded estimates from the five-factor reference model in the empirical example with the parameters for items $51$--$100$ set equal to the parameters for items $1$--$50$. The factor correlation matrix is a $10 \times 10$ block diagonal matrix with rounded FFM estimates on the main-diagonals and zeros on the off-diagonals. We conduct $100$ replications for each $N = \num{625}$, $\num{1250}$, $\num{2500}$, and $\num{5000}$. For I-WAVE, we set $R = 10$ since this value performed well in the previous simulation. We set MH-RM hyperparameters to the mirt package defaults, which performed well across $N$ settings.

\begin{figure}
    \centering
    \begin{subfigure}{0.5\textwidth}
    \includegraphics[width=\textwidth]{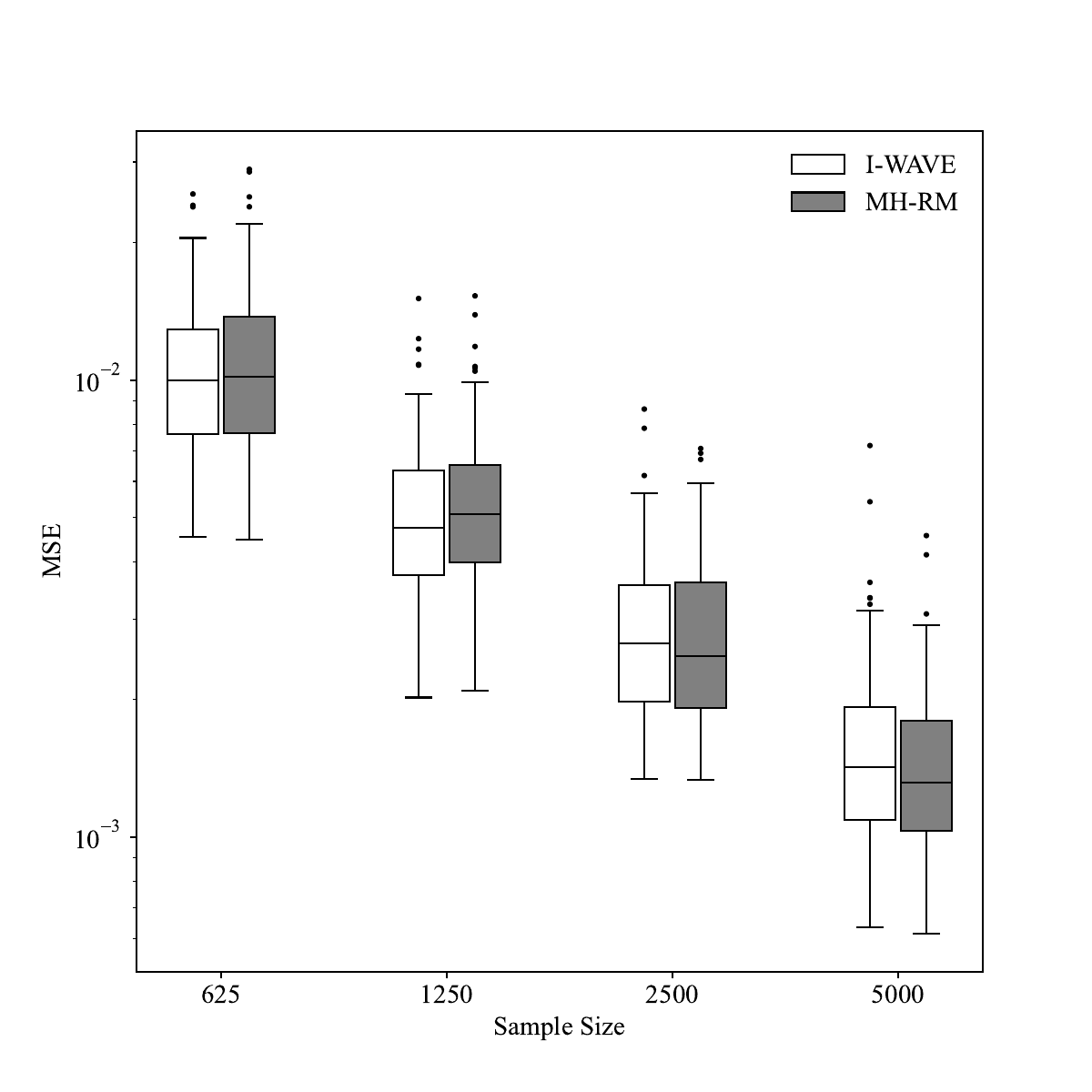}
        \caption{Factor loadings.} \label{fig:6a}
    \end{subfigure}%
    \hfill
    \begin{subfigure}{0.5\textwidth}
        \includegraphics[width=\textwidth]{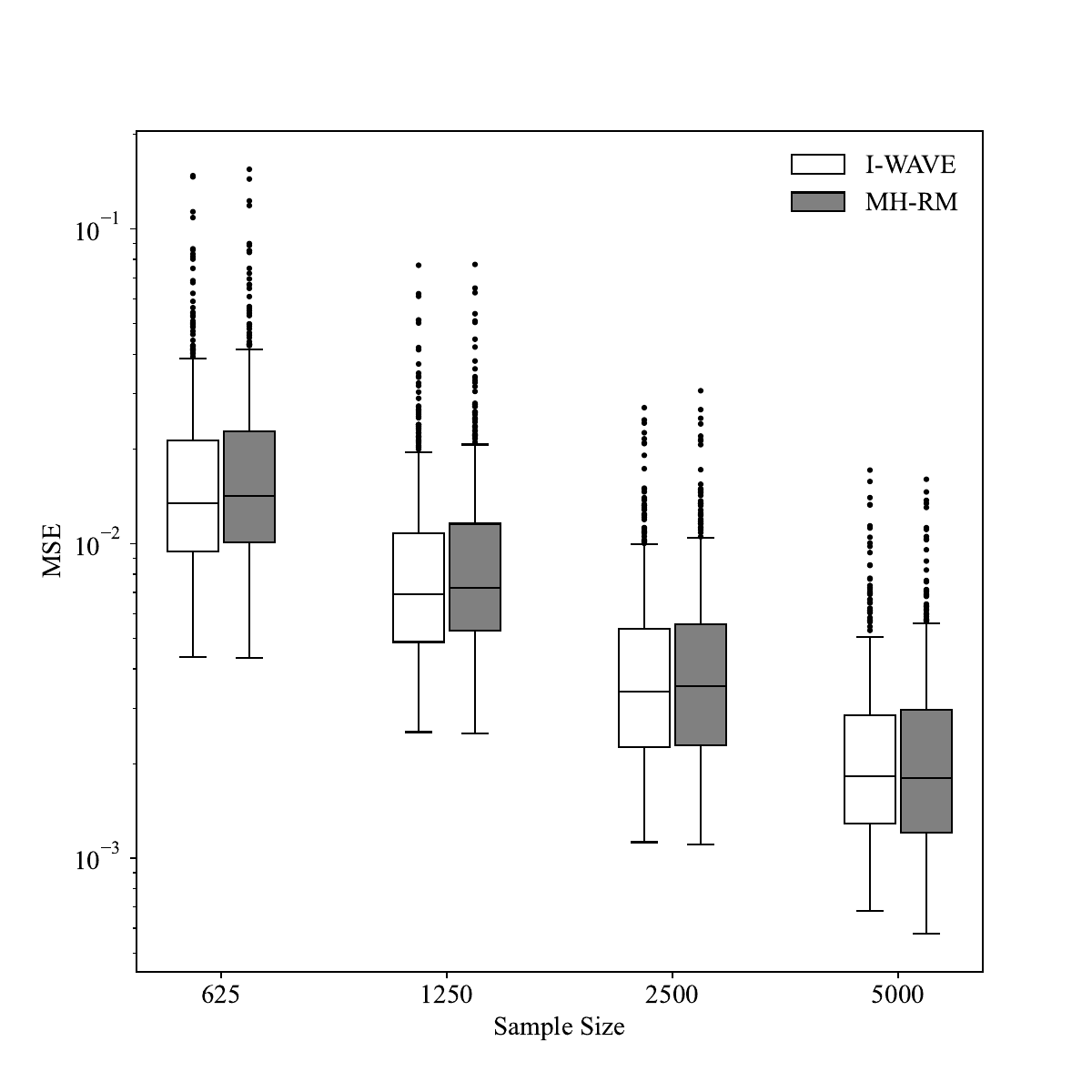}
        \caption{Category intercepts.} \label{fig:6b}
    \end{subfigure}

    \begin{subfigure}{0.5\textwidth}
        \includegraphics[width=\textwidth]{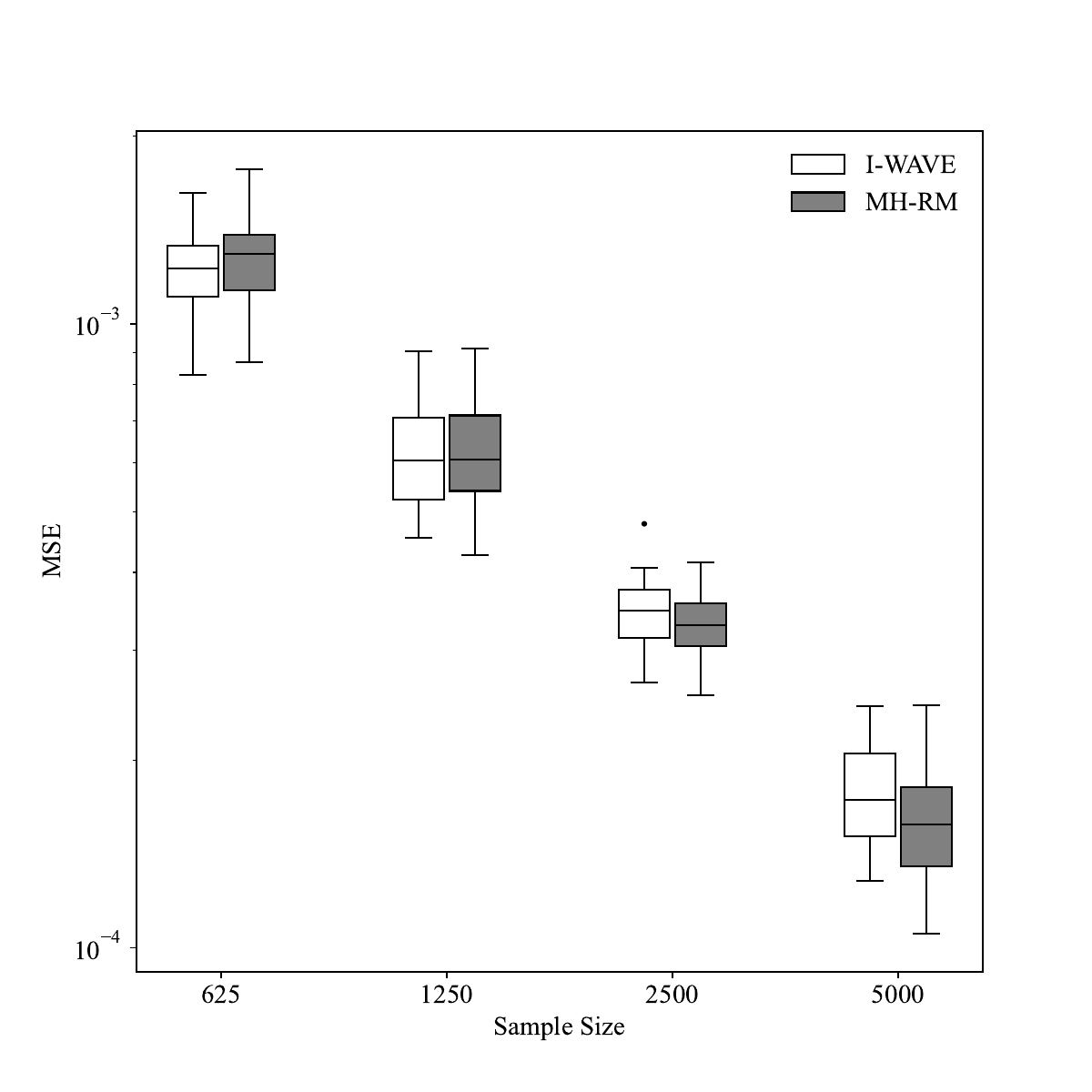}
        \caption{Factor correlations.} \label{fig:6c}
    \end{subfigure}
    \caption{MSE for I-WAVE and the marginal maximum likelihood estimator. MH-RM = Metropolis-Hastings Robbins-Monro.} \label{fig:6}
\end{figure}

\begin{figure}
	\centering
	\includegraphics[width=0.6\textwidth]{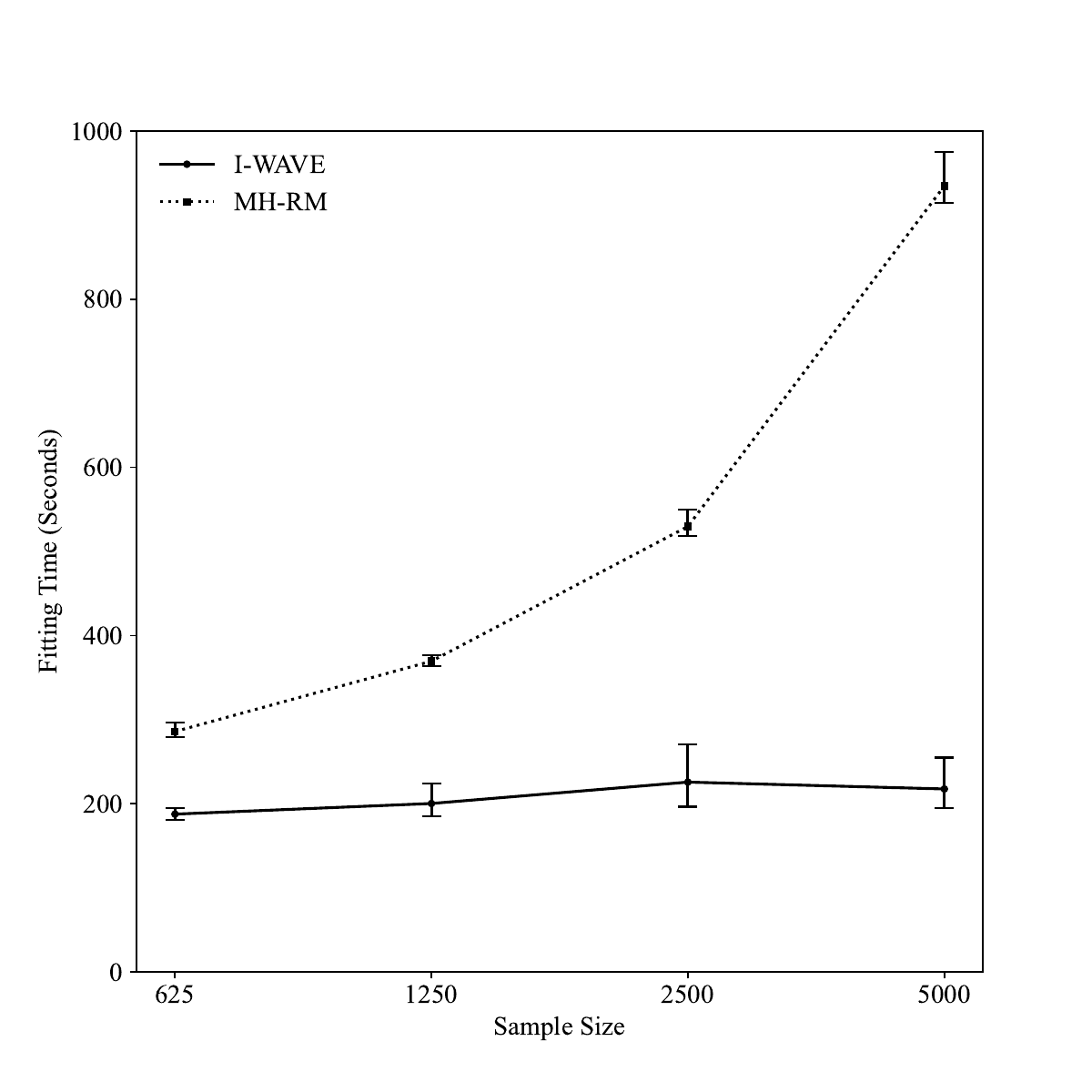}
	\caption{Fitting times for I-WAVE and the marginal maximum likelihood estimator.} \label{fig:7}
\end{figure}

Simulation results are presented in Figures~\ref{fig:6} and~\ref{fig:7}. Both methods obtain comparable estimates in all $N$ settings and have MSE decreasing toward zero with increasing $N$. I-WAVE may obtain slightly better estimates than MH-RM for $N = \num{625}$, and vice versa for $N = \num{5000}$, although the differences appear to be somewhat negligible. Parameter biases for both methods were also comparable and are not shown. I-WAVE is faster than MH-RM in all settings: the median fitting time for I-WAVE remains slightly above three minutes for all $N$, whereas the median fitting time for MH-RM is slightly under $5$ minutes when $N = \num{625}$ and increases to around $15.5$ minutes when $N = \num{5000}$. We note that the MH-RM implementation in the commercially available flexMIRT software \parencite{Cai2017} has core functions written in C++ and is likely faster than mirt, although even this implementation would become slower as $N$ increases.

\subsubsection{Evaluating C2ST-As in a Non-IFA Setting}

It is challenging to verify that C2ST-As have accurate empirical type I error rates and power in the confirmatory IFA setting because precisely controlling the effect size $\varepsilon$ is not straightforward. We instead verify these properties using real-valued data drawn from tractable distributions whereby $\varepsilon$ is precisely controlled. Descriptions and results of these experiments are provided in Appendix \hyperref[appendix:B]{B}. To briefly summarize, we find that empirical type I error rates are well controlled and that empirical power stays close to the values predicted by the formula in Theorem~\ref{theorem:1}.

\subsubsection{Evaluating C2STs and C2ST-RFIs in a Confirmatory IFA Setting} \label{EvalIFASim}

We now investigate the proposed GOF assessment methods' performance in settings where the fitted IFA model is correctly or incorrectly specified. We consider two data generating models for $J = 50$ five-category items: (1) the same FFM considered in the previous simulation studies as well as (2) an SFM with five correlated factors each measured by $10$ items, one orthogonal doublet factor measured by items $17$ and $18$, and another orthogonal doublet factor measured by items $41$ and $48$. Generating parameters for (2) are rounded estimates from the seven-factor reference model in the empirical example. We simulate $100$ data sets from (1) and (2) for each $N = \num{625}$, $\num{1250}$, $\num{2500}$, $\num{5000}$, and $\num{10000}$. Using I-WAVE with the same hyperparameters as in the empirical example, we fit two models to each data set: (a) an FFM with the same specification as (1) and (b) an SFM with the same specification as (2). This leads to four different settings for each combination of data generating model and fitted model. Viable GOF assessment methods should indicate near perfect fit for settings (1a) and (2b), which have correctly specified fitted models, as well as for setting (1b), which has an overspecified fitted model. Setting (2a), on the other hand, has an underspecified fitted model and should demonstrate poorer fit as $N$ increases. Although it might be expected that $\acc = 0.5$ in settings with correctly specified and overspecified models, it is also feasible that IFA model parameter estimate uncertainty leads to $\acc$ slightly greater than $0.5$. To investigate both possibilities, we assess GOF for all simulation settings using exact C2STs where $\delta = 0$ as well as C2ST-As where $\delta = 0.025$, which respectively correspond to $H_0 : \acc = 0.5$ and $H_0 : \acc = 0.525$.

\begin{figure}
    \centering
    \begin{subfigure}{\textwidth}
    \includegraphics[width=\textwidth]{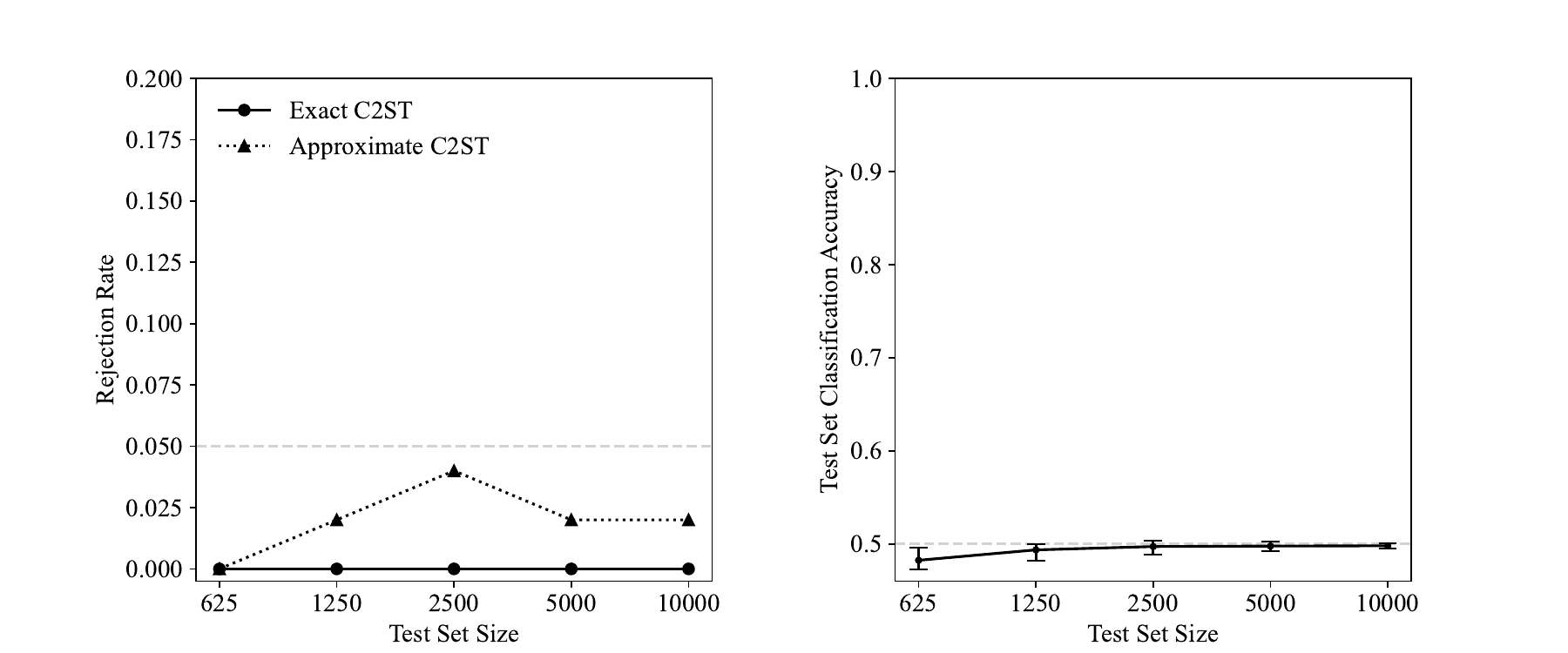}
        \caption{An overspecified seven-factor fitted model.} \label{fig:8a}
    \end{subfigure}
    
    \begin{subfigure}{\textwidth}
        \includegraphics[width=\textwidth]{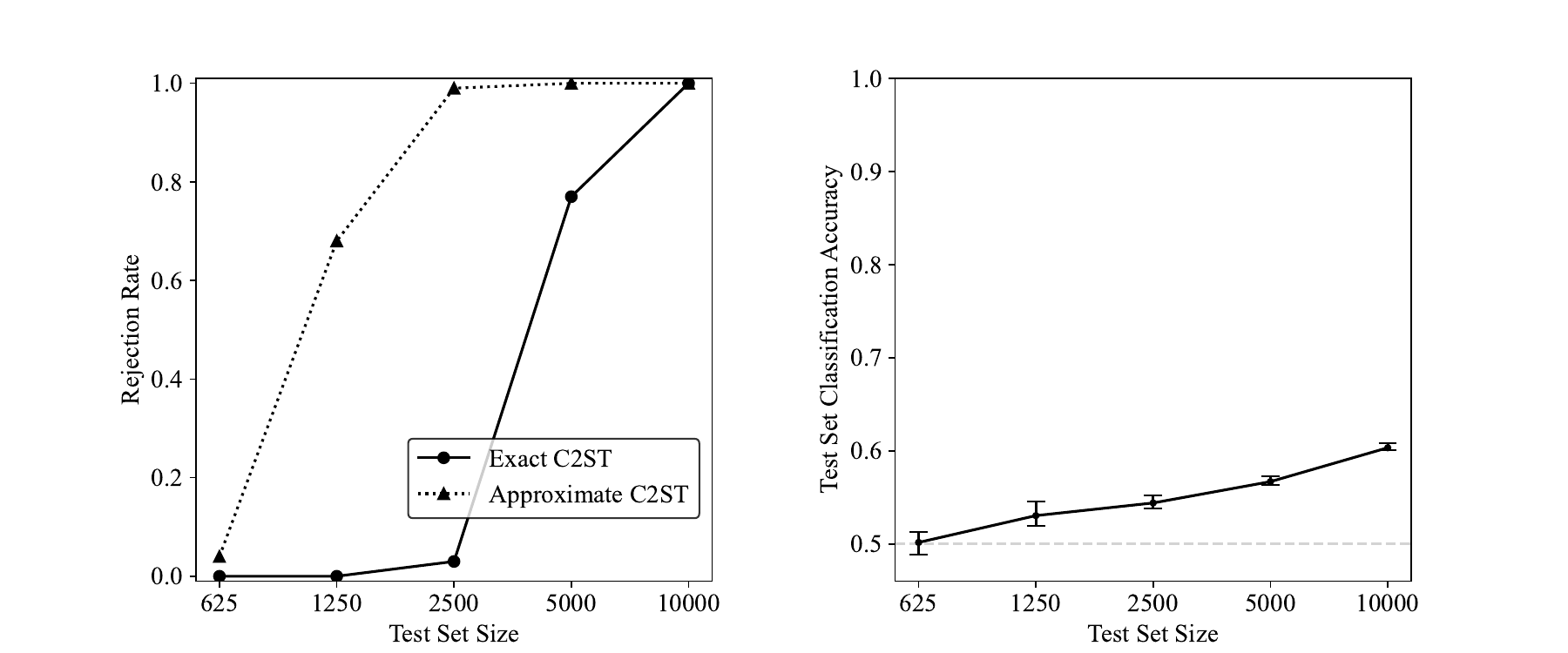}
        \caption{An underspecified five-factor fitted model.} \label{fig:8b}
    \end{subfigure}
    \caption{Rejection rates as well as test set classification accuracies for approximate and exact C2STs.} \label{fig:8}
\end{figure}

Rejection rates at signficance level $\alpha = 0.05$ as well as test set classification accuracies for the overspecified SFM in setting (1b) are shown in Figure~\ref{fig:8a}. Results for the correctly specified models in settings (1a) and (2b) were nearly identical and are not shown. Rejection rates remain well below the nominal level and test set classification accuracies come close to $0.5$ as $N_{\test}$ increases, suggesting that the SFM fits the simulated data nearly perfectly. The near-zero rejection rates (as opposed to rejection rates near $\alpha = 0.05$) likely occurred because there was almost no signal in the training data, leading classifiers to overfit to noise and perform worse than random chance. We emphasize that this finding does not imply that C2STs are conservative tests --- indeed, results in Appendix \hyperref[appendix:B]{B} suggest that in addition to attaining empirical power close to theoretically predicted values, C2STs maintain the nominal level when $H_0$ is true and there is signal in the training data.

\begin{figure}
	\centering
	\includegraphics[width=0.8\textwidth]{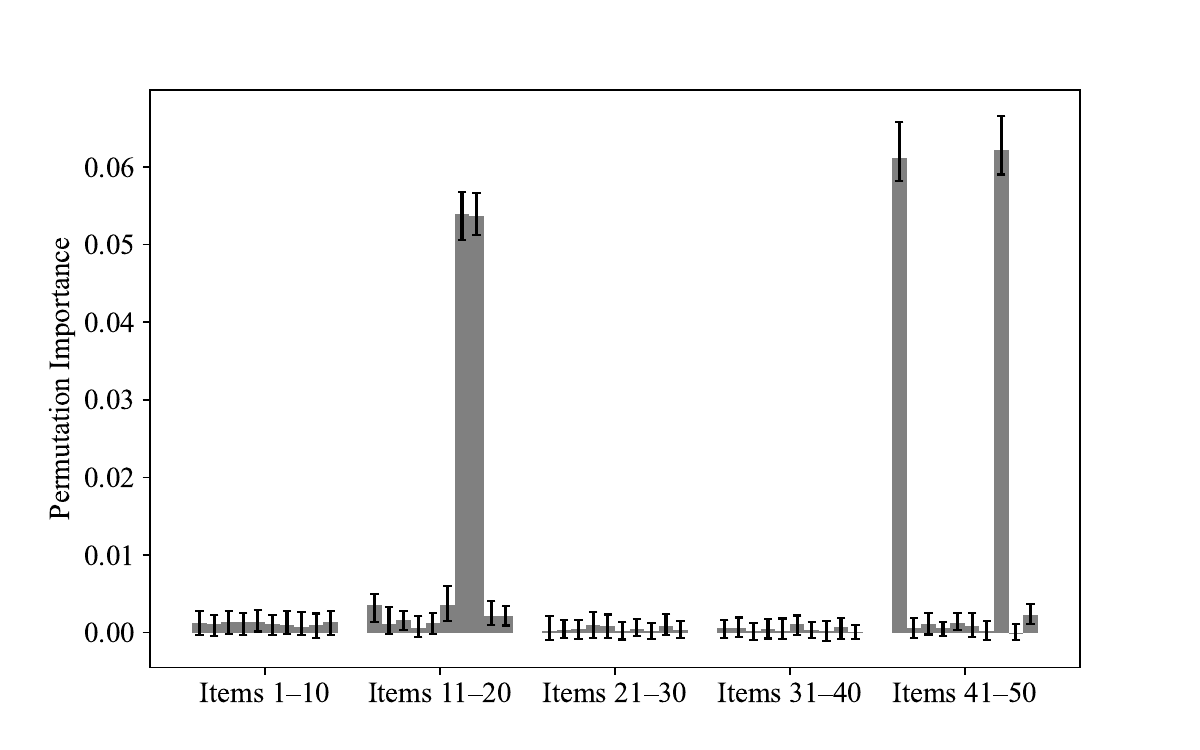}
	\caption{Permutation importances for the underspecified five-factor fitted model when $N = \num{10000}$.} \label{fig:9}
\end{figure}

C2ST results for the underspecified fitted model in setting (2a) are shown in Figure~\ref{fig:8b}. NN classifiers performed well, with classification accuracies exceeding $0.5$ and rejection rates tending to one as $N_{\test}$ increases. Rejection rates for C2ST-As tend to one more slowly than for exact C2STs, which is the intended behavior and shows that C2ST-As are relatively tolerant of model misspecification compared to exact C2STs. We also demonstrate the viability of PIs by assessing the underspecified fitted model's item-level fit when $N = \num{10000}$. Results in Figure~\ref{fig:9} show that NNs clearly flagged items $17$, $18$, $41$, and $48$ as poor fitting with all other items having PIs near zero.

\begin{figure}
	\centering
	\includegraphics[width=0.6\textwidth]{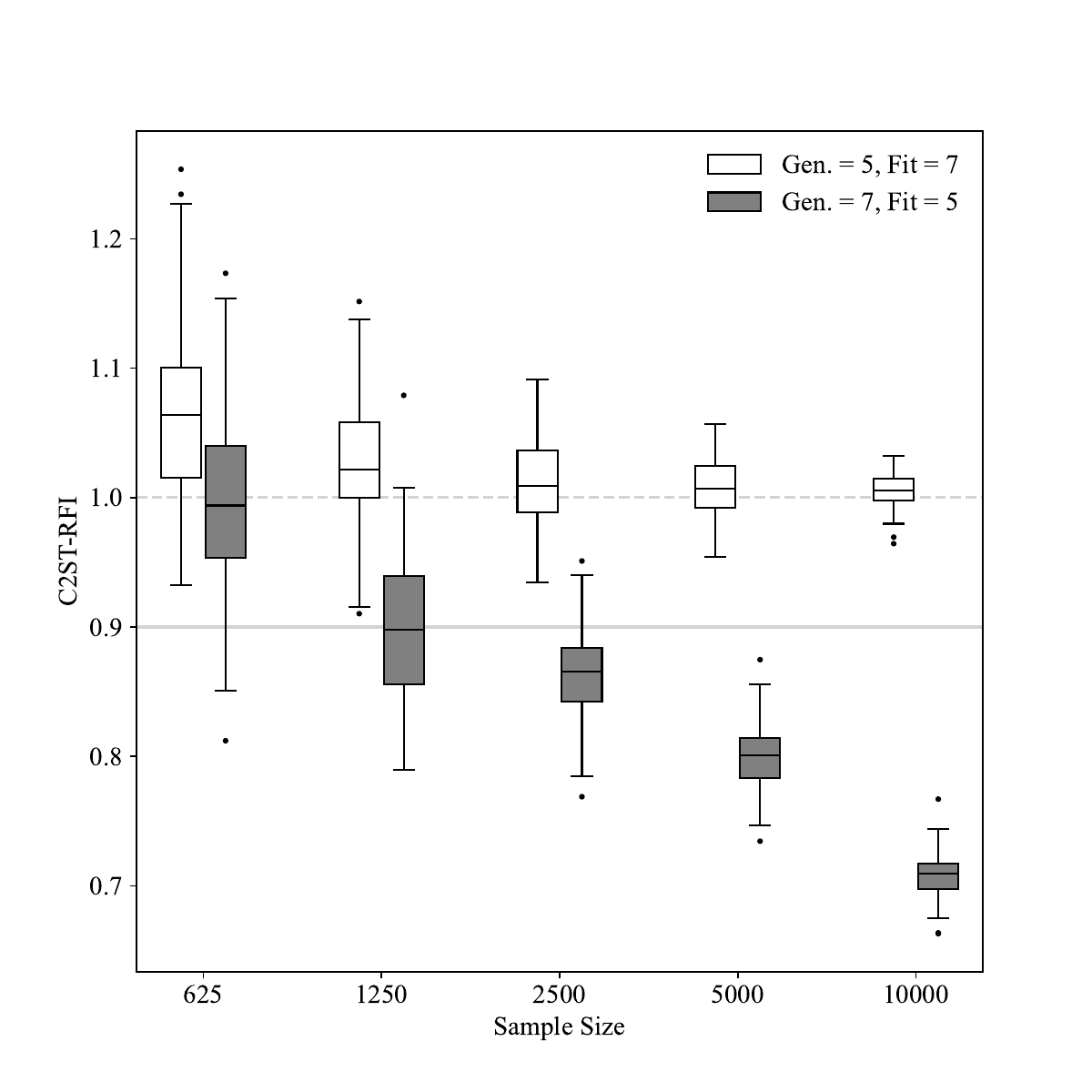}
	\caption{C2ST-RFIs for the overspecified seven-factor fitted model ($\text{Gen.} = 5$, $\text{Fit} = 7$) and for the underspecified five-factor fitted model ($\text{Gen.} = 7$, $\text{Fit} = 5$). The provisory cutoff of $0.9$ is marked with a solid line.} \label{fig:10}
\end{figure}

Boxplots of C2ST-RFIs for the overspecified SFM in setting (1b) as well as for the underspecified FFM in setting (2a) are shown in Figure \ref{fig:10}. C2ST-RFIs for the correctly specified models in settings (1a) and (2b) were nearly identical to C2ST-RFIs for (1b) and are not shown. The means of the sampling distributions of C2ST-RFIs appear to depend on $N$. C2ST-RFIs for both (1b) and (2a) start out larger than one when $N_{\test} = 625$, with RFIs for (1b) tending to one and RFIs for (2a) tending to values smaller than one as $N_{\test}$ increases. These results suggest that the SFM fit its data almost perfectly and that the FFM fit its data relatively poorly, although classifiers overfitted to noise in the training data for small $N_{\train}$. The provisory cutoff of $0.9$ obtains similar rejection rates to those obtained by C2STs: essentially no C2ST-RFIs fall below the threshold in setting (1a) (i.e., the rejection rate remains near zero) and all C2ST-RFIs in setting (2b) fall below the threshold as $N_{\test}$ increases (i.e., the rejection rate tends to one).

\begin{figure}
    \centering
    \begin{subfigure}{0.6\textwidth}
    \includegraphics[width=\textwidth]{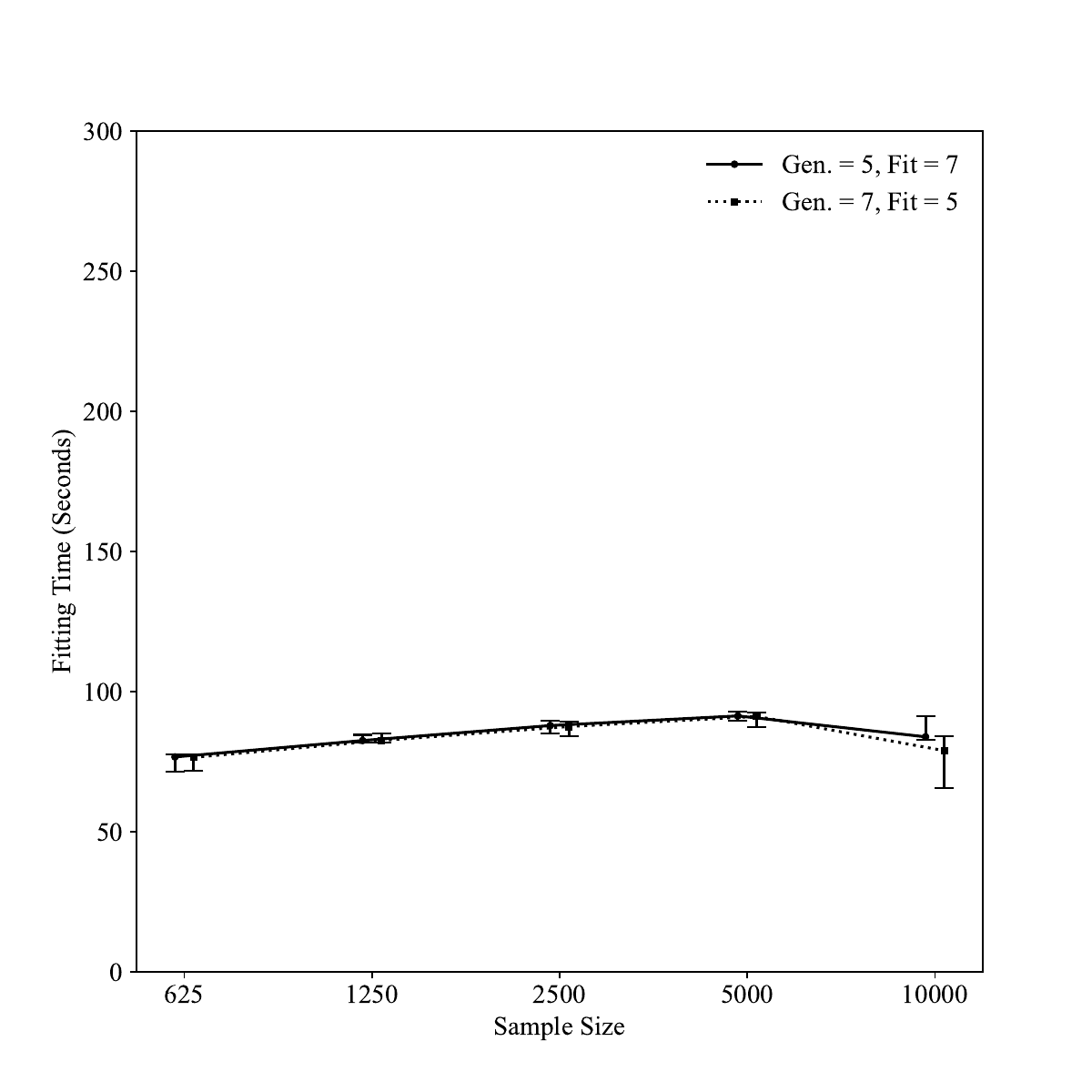}
        \caption{C2ST run times.} \label{fig:11a}
    \end{subfigure}
    \begin{subfigure}{0.6\textwidth}
        \includegraphics[width=\textwidth]{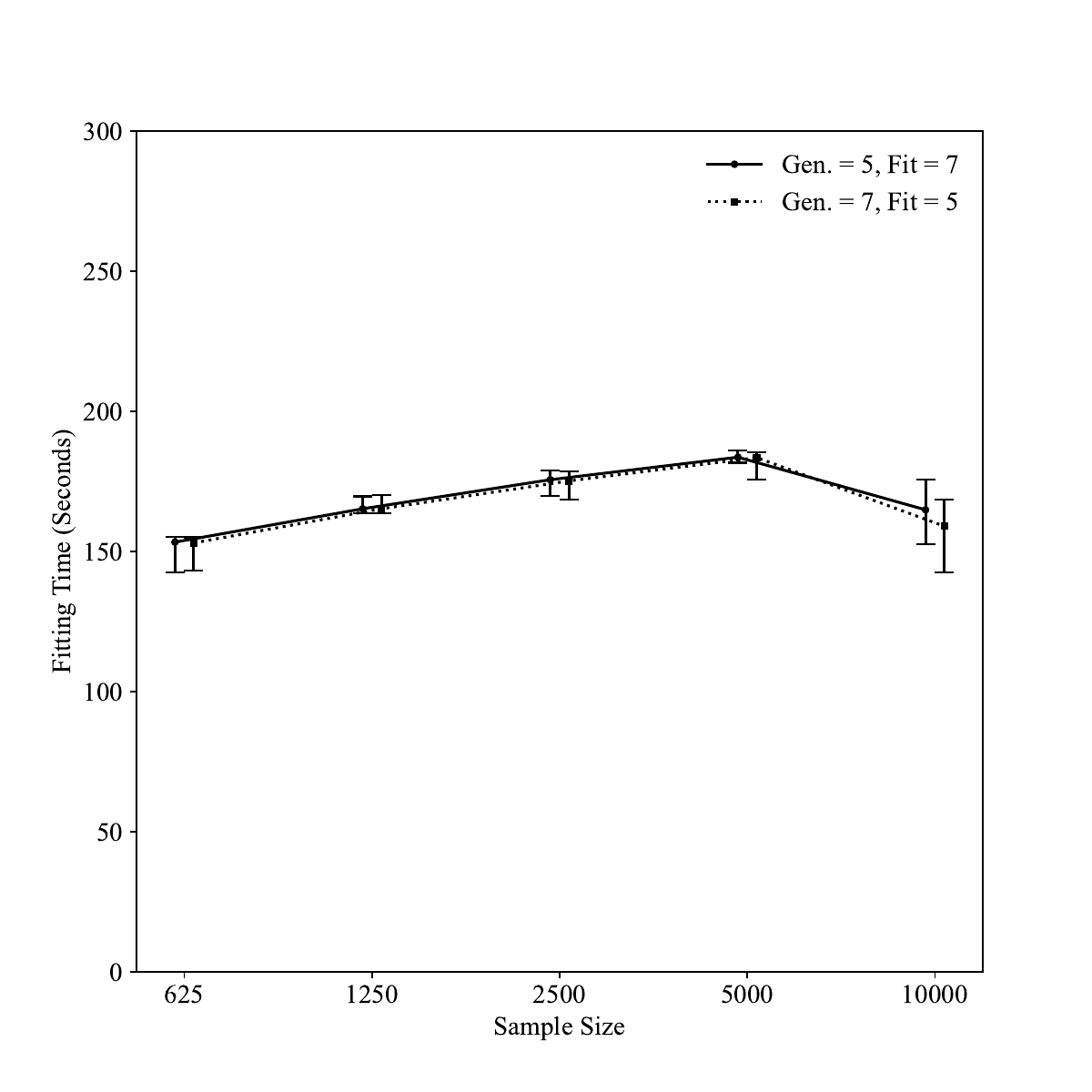}
        \caption{C2ST-RFI run times.} \label{fig:11b}
    \end{subfigure}
    \caption{Run times for C2STs and C2ST-RFIs from the simulation settings with the overspecified seven-factor fitted model and the underspecified five-factor fitted model.} \label{fig:11}
\end{figure}

We assessed each GOF assessment method's computational efficiency by computing the total time required to sample all synthetic data and to fit NN classifiers. Total run times for C2STs and C2ST-RFIs computed in simulation setting (2a) are shown in Figures~\ref{fig:11a} and~\ref{fig:11b}, respectively. Run times for other simulation settings were very similar and are not shown. Run times remain close to constant as $N_{\test}$ increases, with median run times for C2STs and C2ST-RFIs staying around $84$ and $168$, respectively. These findings demonstrate that C2STs and C2ST-RFIs are viable GOF assessment methods even in the large $N$ setting.

\section{Discussion} \label{section:5}

This work is concerned with the theoretical properties and empirical performance of machine learning-based parameter estimation and goodness-of-fit assessment methods for large-scale confirmatory item factor analysis. An importance-weighted amortized variational estimator implemented via a deep learning algorithm demonstrated improved parameter recovery as the sample size increased and obtained comparable estimates to those obtained by the MML estimator implemented via the MH-RM algorithm. The deep learning algorithm's computational efficiency appears to be essentially independent of the sample size, enabling fitting even with extremely large samples. A simulation-based test of exact fit called the classifier two-sample test was described and extended into a test of approximate fit as well a relative fit index. Approximate C2STs and C2ST-RFIs successfully identified when an IFA model was correctly or incorrectly specified. A permutation importance technique was demonstrated as a tool for exploring piece-wise model fit.

The proposed methods have a number of limitations and extensions that may be addressed in future work.

First, computing standard errors (SEs) and handling missing data with I-WAVE were not discussed. Approximate SEs may be obtained by evaluating the observed information matrix at the parameter estimates obtained by maximizing the IW-ELBO, then inverting this matrix block-wise \parencite{Hui2017}. As noted by \textcite{UrbanBauer2021}, SEs will likely be small for the large-scale applications considered here. Procedures for handling missing-at-random data \parencite{MatteiMIWAE2019} as well as missing-not-at-random data \parencite{Ipsen2021} have been developed for general amortized importance-weighted VI and can be straightforwardly utilized for I-WAVE.

Second, using approximate C2STs to assess person and piece-wise fit may be more thoroughly investigated. In terms of person fit, the approach suggested in Sect. \ref{ExactC2STs} remains to be investigated via simulation studies. In terms of piece-wise fit, the permutation importance measures applied here have shortcomings including (1) unknown sampling distributions and (2) degraded performance when predictors are highly correlated \parencite[e.g.,][]{Hooker2019}. For (1), future work may explore whether accurate $p$-values and confidence intervals for estimated PIs can be obtained via parametric approximations \parencite{Altmann2010} or bootstrap resampling. For (2), it could be fruitful to explore alternative approaches based on Shapley additive explanations \parencite[SHAP;][]{Lundberg2017}, a game theoretic method for explaining fitted classifiers' predictions with variants that perform well in the presence of multicollinearity \parencite{Aas2021, Basu2020, Sellereite2020}.

Third, C2ST-RFIs may be improved by developing better (1) cutoff criteria and (2) model complexity penalties. For (1), although our provisory cutoff criterion of $0.9$ performed well in a narrow set of conditions, further research is needed to develop more robust cutoff criteria for evaluating model fit in practice. For (2), we employed a complexity penalty based on the number of fitted parameters. Although this parameter counting approach is fast, alternative penalties that take into account IFA models' functional forms would provide more accurate characterizations of complexity that could be used to compare different models with the same number of parameters \parencite[e.g.,][]{Bonifay2017}.

In summary, the methods considered in this work may provide feasible and promising frameworks for testing hypotheses about the latent structure underlying large-scale item response data in a computationally efficient manner. Both I-WAVE and C2STs are highly flexible frameworks that may be extended in a variety of ways. Some of these extensions are discussed above, some are discussed in the vast machine learning literature, and some are yet to be conceived. We view this work as part of a dialogue between machine learning and psychometrics that is leading to the development of new extensions and applications with the potential to positively impact both fields.

\setcounter{secnumdepth}{0} 

\printbibliography

\appendix

\section{Proof of Theorem 1} \label{appendix:A}

At significance level $\alpha$, the approximate decision threshold for $\acc$ is
\begin{equation} \nonumber
    z_{\alpha} = \frac{1}{2} + \delta + \sqrt{\frac{\frac{1}{4} - \delta^2}{N_{\test}}} \Phi^{-1}(1 - \alpha).
\end{equation}
When $\acc < z_{\alpha}$, we accept $H_0$. The approximate probability of making a type II error (i.e., of incorrectly accepting $H_0$ when $H_0$ is false) is
\begin{align}
    \Pr_{\mathcal{N}\left(\acc \,\middle\vert\, \frac{1}{2} + \delta + \varepsilon, \frac{\frac{1}{4} - \delta^2 - 2 \delta \varepsilon - \varepsilon^2}{N_{\test}} \right)}(\acc < z_{\alpha}) &= \Pr_{\mathcal{N}\left(\acc' \,\middle\vert\, 0, \frac{\frac{1}{4} - \delta^2 - 2 \delta \varepsilon - \varepsilon^2}{N_{\test}} \right)} \left( \acc' < \sqrt{\frac{\frac{1}{4} - \delta^2}{N_{\test}}} \Phi^{-1}(1 - \alpha) - \varepsilon \right) \nonumber \\
    &= \Phi \left(  \sqrt{\frac{N_{\test}}{\frac{1}{4} - \delta^2 - 2 \delta \varepsilon - \varepsilon^2}} \left( \sqrt{\frac{\frac{1}{4} - \delta^2}{N_{\test}}} \Phi^{-1}(1 - \alpha) - \varepsilon \right) \right) \nonumber \\
    &= \Phi \left( \frac{\sqrt{\frac{1}{4} - \delta^2} \Phi^{-1}(1 - \alpha) - \varepsilon \sqrt{N_{\test}}}{\sqrt{\frac{1}{4} - \delta^2 - 2 \delta \varepsilon - \varepsilon^2}} \right). \nonumber
\end{align}
The power of the C2ST-A is therefore approximately given by
\begin{align}
    \power(\alpha, N_{\test}, \delta, \varepsilon) &\approx 1 - \Phi \left( \frac{\sqrt{\frac{1}{4} - \delta^2} \Phi^{-1}(1 - \alpha) - \varepsilon \sqrt{N_{\test}}}{\sqrt{\frac{1}{4} - \delta^2 - 2 \delta \varepsilon - \varepsilon^2}} \right) \nonumber \\
    &= \Phi \left( \frac{\varepsilon \sqrt{N_{\test}} - \sqrt{\frac{1}{4} - \delta^2} \Phi^{-1}(1 - \alpha)}{\sqrt{\frac{1}{4} - \delta^2 - 2 \delta \varepsilon - \varepsilon^2}} \right). \nonumber
\end{align}
\qed

\vspace{\fill}\pagebreak

\section{Type I Error Rates and Power for C2ST-As} \label{appendix:B}

We conduct a small simulation study to verify that C2ST-As have empirical type I error close to $\alpha = 0.05$ as well as empirical power close to values predicted by the formula in Theorem \ref{theorem:1}. Type I error rates for exact C2STs were empirically verified by \textcite{Lopez-Paz2017} and are not considered here. Since it is difficult to control the effect size $\varepsilon$ in the IFA setting, we instead consider real-valued data for this study.

\begin{figure}
	\centering
	\includegraphics[width=\textwidth]{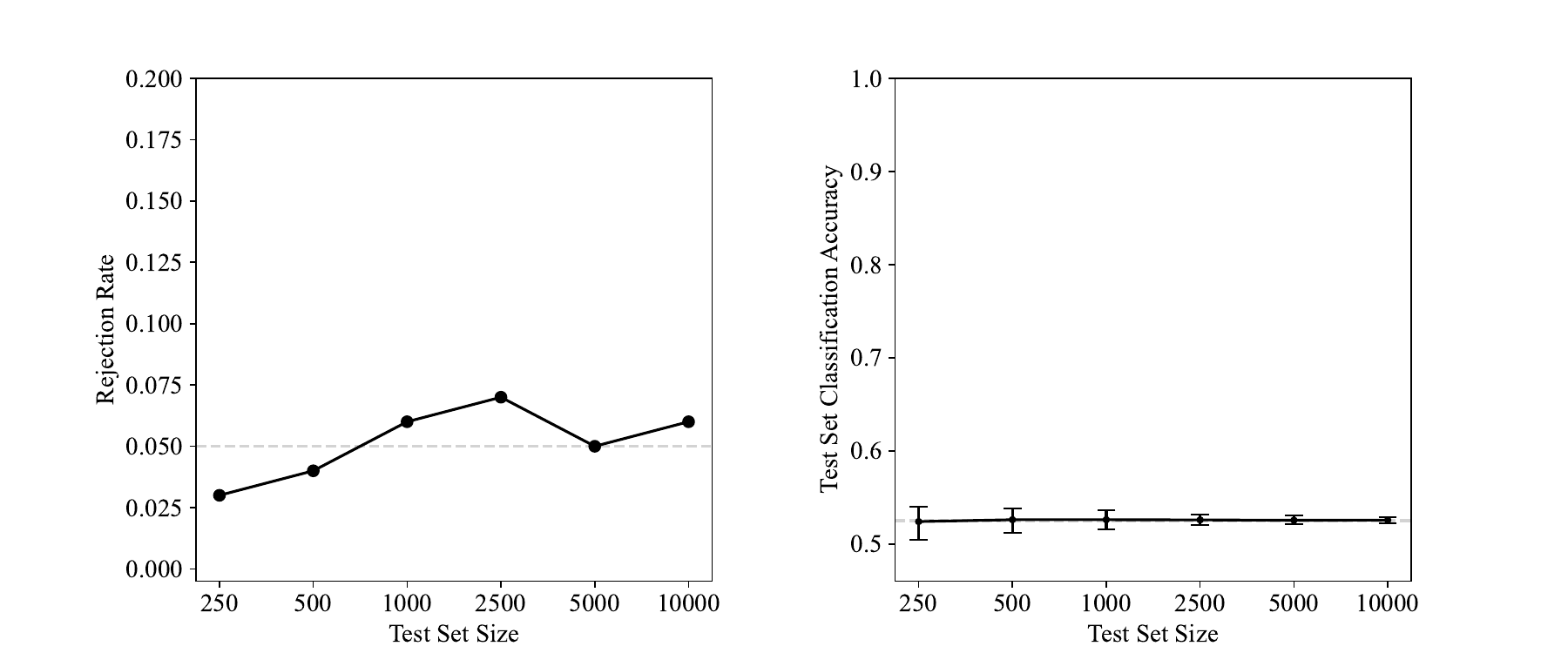}
	\caption{Empirical type I error and test set classification accuracy for C2ST-As.} \label{fig:A2}
\end{figure}

We begin with the type I error experiments. For each replication $a = 1, \ldots, 100$, we simulate data by drawing two samples $\{ x_i^{(a)} \}_{i = 1}^N \sim \mathbb{P} = \mathcal{U}(x_i \mid 0, 1)$ and $\{ y_i^{(a)} \}_{i = 1}^N \sim \hat{\mathbb{P}} = \mathcal{U}(y_i \mid 0.05, 1.05)$ where $N = 250$, $500$, $\num{1000}$, $\num{2500}$, $\num{5000}$, and $\num{10000}$. $\mathbb{P}$ and $\hat{\mathbb{P}}$ are shown in Figure~\ref{fig:A1}. Close to $95\%$ of the observations in each data set will fall in the region where $\mathbb{P}$ and $\hat{\mathbb{P}}$ overlap (i.e., in $[0.1, 1]$) and will be indistinguishable since $\mathbb{P} = \hat{\mathbb{P}}$. On the other hand, close to $5\%$ of the observations will fall either to the left or the right of the overlapping region (i.e., either in $[0, 0.1)$ or $(1, 1.05]$, respectively) and will be perfectly distinguishable since $\hat{\mathbb{P}} = 0$ to the left and $\mathbb{P} = 0$ to the right. This implies that the maximum obtainable test set classification accuracy is close to $0.525$. We therefore test $H_0 : \acc = 0.525$ against $H_1 : \acc > 0.525$ so that the effect size $\varepsilon = 0$. Figure~\ref{fig:A2} shows that C2ST-As have rejection rates close to the nominal level in all $N$ settings and test set classification accuracies converging to $0.525$ as $N_{\test}$ increases.

\begin{figure}
	\centering
    \resizebox{0.6\textwidth}{!}{
        \begin{tikzpicture}
        \begin{axis}[
            samples = 100,
            const plot mark mid,
            scale only axis=true,
            ymin = 0, ymax = 3,
            ytick = {0, 2},
            yticklabels = {0, 1},
            xmin = -0.5, xmax = 2.6,
            xtick = {0, 2},
            xticklabels = {0, 1},
            tick align = outside,
            every tick/.style={
                black,
            },
            xtick pos=left,
            ytick pos=left,
            legend style={draw=none},
            legend cell align={left},
            xlabel = {$x$},
            ylabel = {$p(x)$}]
        \addplot+[very thick, mark = none, const plot,
                  color = black] coordinates
            {(0,0) (0,2) (2,2) (2,0)};
        \addplot+[very thick, mark = none, const plot,
                  color = black, densely dashed] coordinates
            {(0.1,0) (0.1,2) (2.1,2) (2.1,0)};
        \addplot[name path = top1, domain = 0:0.1,
                 samples = 100] {2};
        \addplot[name path = axis1, domain = 0:0.1,
                 samples = 100] {0};
        \addplot[name path = top2, domain = 2:2.1,
                 samples = 100] {2};
        \addplot[name path = axis2, domain = 2:2.1,
                 samples = 100] {0};
        \addplot[name path = top3, domain = 0.1:2,
                 samples = 100] {2};
        \addplot[name path = axis3, domain = 0.1:2,
                 samples = 100] {0};
         \addplot [fill = gray,  fill opacity=0.75]
            fill between[of = top1 and axis1, soft clip={domain=0:0.1},];
         \addplot [fill = gray,  fill opacity=0.75]
            fill between[of = top2 and axis2, soft clip={domain=2:2.1},];
         \addplot [fill = gray, fill opacity=0.25]
            fill between[of = top3 and axis3, soft clip={domain=0.1:2},];
        \addlegendentry{$\mathbb{P} = \mathcal{U}(x \mid 0, 1)$}
        \addlegendentry{$\hat{\mathbb{P}} = \mathcal{U}(x \mid 0.05, 1.05)$}
        \end{axis}
        \end{tikzpicture}
    }
    \caption{Uniform generating distributions used to verify the type I error rate for approximate C2STs. Observations falling in the overlapping (light gray) region are indistinguishable, while observations falling in the non-overlapping (dark gray) regions are perfectly distinguishable.} \label{fig:A1}
\end{figure}

\begin{figure}
	\centering
	\includegraphics[width=\textwidth]{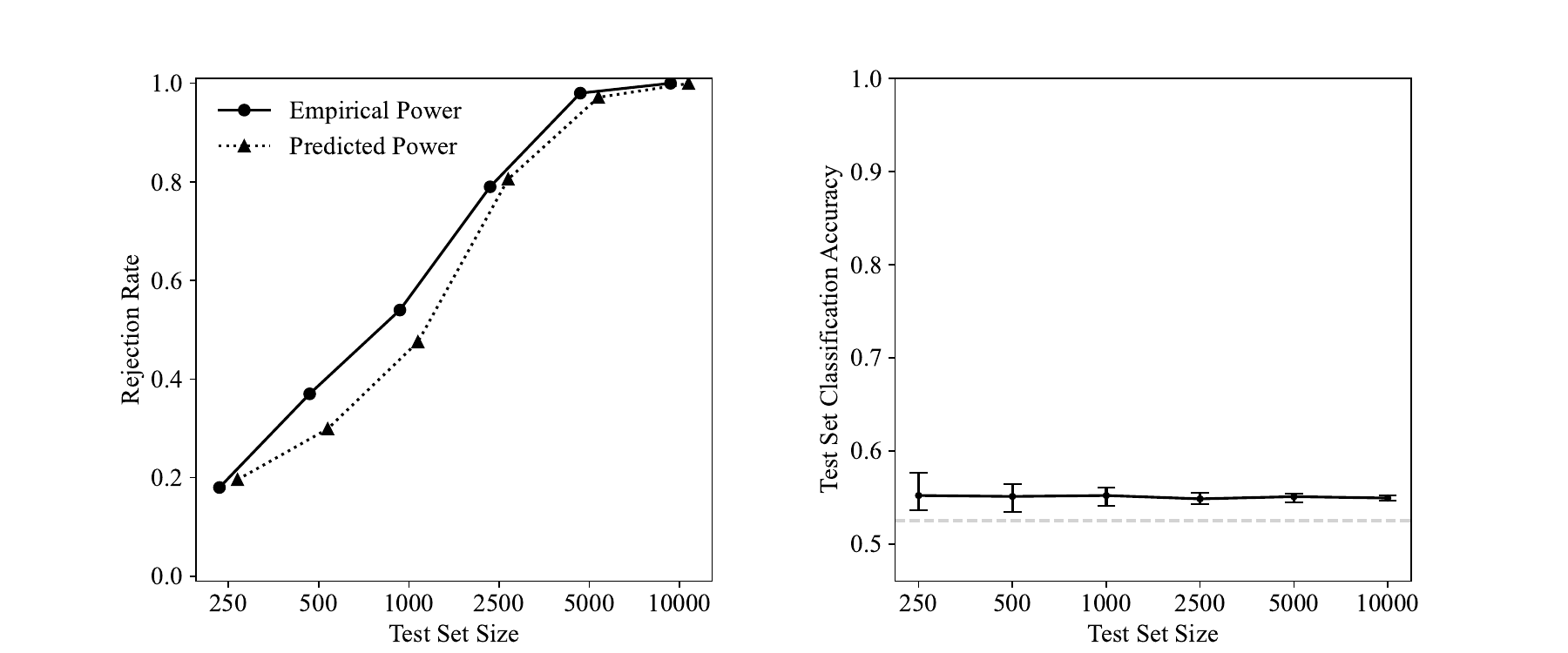}
	\caption{Empirical power and and test set classification accuracy for C2ST-As.} \label{fig:A3}
\end{figure}

To assess empirical power, we follow the same procedure as above except we now draw $\{ y_i^{(a)} \}_{i = 1}^N \sim \hat{\mathbb{P}} = \mathcal{U}(y_i \mid 0.1, 1.1)$. By similar reasoning to that given above, the maximum obtainable test set classification accuracy in this setting is close to $0.55$. We again test $H_0 : \acc = 0.525$ against $H_1 : \acc > 0.525$ so that now the effect size $\varepsilon = 0.025$. Figure~\ref{fig:A3} shows that C2ST-As have both power converging to one and test set classification accuracies converging to $0.55$ as $N_{\test}$ increases. Power values predicted by the formula in Theorem~\ref{theorem:1} are close to the empirical power values for all tests.

\end{document}